\documentclass[10pt,letterpaper]{article}

\usepackage[T1]{fontenc}
\usepackage[utf8]{inputenc}
\usepackage{microtype}
\usepackage{graphicx}
\usepackage[section]{placeins}
\usepackage{booktabs,tabularx,array}
\usepackage{amsmath,amssymb}
\usepackage[table]{xcolor}
\usepackage{enumitem}
\usepackage[numbers,sort&compress]{natbib}
\usepackage[preprint,logo,nonatbib]{agibot_enerverse}
\hypersetup{pdftitle={PhysiOmniDuplex-Bench},pdfauthor={}}
\graphicspath{{figures/}}
\setlist[itemize]{nosep,leftmargin=*}
\setlist[enumerate]{nosep,leftmargin=*}
\newcommand{\benchmark}{PhysiOmniDuplex-Bench}

\newcolumntype{Y}{>{\raggedright\arraybackslash}X}

\DeclareRobustCommand{\CorrespondingEnvelope}{%
  \setlength{\unitlength}{0.65pt}%
  \begin{picture}(11,8)
  \put(0,0){\framebox(11,8){}}%
  \qbezier(0,8)(2.75,5.75)(5.5,3.5)%
  \qbezier(5.5,3.5)(8.25,5.75)(11,8)%
  \end{picture}%
}

\title{PhysiOmniDuplex-Bench:\\
A Benchmark for Embodied Interactive Agents\\
in Real-World Physical Environments}
\author{%
\textbf{Kang Feng}\textsuperscript{\CorrespondingEnvelope}\quad \textbf{Yi Wang}\quad \textbf{Jian Wang}\quad \textbf{Lin Shen}\quad \textbf{Dong Liu}\quad \textbf{Weijie Wang}\\[2pt]
\textbf{Mingcai Yuan}\quad \textbf{Chang Xu}\quad \textbf{Shaobo Wang}\quad \textbf{Huijun Zhang}\quad \textbf{Shaolei Yang}\quad \textbf{Xiaoye Qu}\\[2pt]
\textbf{Zhijun Tu}\quad \textbf{Anna Wang}\quad \textbf{Shengqiang Liu}\quad \textbf{Rachel}\textsuperscript{\CorrespondingEnvelope}\\[5pt]
AGIBOT\\[2pt]
{\small\ttfamily\color{blue!65!black}\{fengkang, rachel\}@agibot.com}}
\date{}
\begin{document}
\maketitle
\begin{center}
\end{center}

\begin{abstract}
As embodied agents take on customer-facing roles in retail, food service, and reception, evaluating how reliably they interact in shared physical environments becomes a practical requirement for real-world applications. The evidence needed to interpret a request is distributed across people, speech, objects, and actions, and changes as the interaction unfolds. The agent’s own actions shape how people respond, while local rules and social boundaries constrain how it can achieve a goal. Evaluating these systems therefore requires evidence about both how an interaction unfolds and what it accomplishes. We introduce \textbf{PhysiOmniDuplex-Bench}, a benchmark built around interaction challenges observed in current applications or anticipated in future service scenarios, organized into four task categories: omnimodal perception, proactive service, full-duplex interaction, and task-oriented interaction. The final evaluation dataset comprises 1,507 interaction sessions, drawn from robot sessions collected with full participant consent, de-identified robot deployment recordings, recordings derived from a public dataset, actor-enacted recordings based on actual user requests, and model-generated audiovisual material. We compare five real-time multimodal services using controlled audiovisual replay and task-specific protocols, scoring their observable responses. A responsive closed-loop protocol is also specified for settings in which people and task state change after the system acts. Scoring  additionally assesses whether an agent can adapt the requested means and still achieve a permissible goal under contextual constraints. The evaluation reveals complementary strengths in perception, turn timing, and task follow-through, but across the evaluated capabilities, no tested configuration performed consistently well throughout complex service interactions.

\end{abstract}

\section{Introduction}
\label{sec:introduction}

As robots take on customer-facing roles in retail, food service, and reception,
their reliability in shared physical environments has become a practical
requirement rather than a research aspiration. This need is further
highlighted by application reports and human--robot interaction studies
in logistics, mall information services, and physical
collaboration \cite{goodrich2007hri,gxo2024digit,yano2026nonverbal,okafuji2022mall,allgeuer2024chatty,lee2025shrec}.
We use \emph{embodied interactive agent} to refer to the configured system
under evaluation, including its model, memory, and available sensing,
interaction, and execution interfaces.

We construct scenarios around robots' current and anticipated service tasks.
Scenarios for current applications draw on observed interactions and the
problems they reveal, while prospective scenarios are designed around
expected duties, site conditions, and foreseeable interaction challenges.
We then translate these scenarios into evaluation cases with explicit
observations, behavioral requirements, and success criteria.

Evaluating embodied interactive agents poses three challenges. First,
the evidence needed to interpret a request is distributed across speakers,
gestures, objects, and spatial relations, and changes as people move or
revise their requests. In a restaurant, a customer may point to identify
whose order they want to copy. Second, the agent's speech and actions
shape subsequent interaction. A person can decline assistance only after
an offer, so systems may encounter different numbers of refusal events.
Third, task success depends on contextual constraints, including
permissions, venue rules, and interpersonal boundaries.

Existing evaluation efforts span multimodal understanding, real-time
interaction, and embodied tasks. Multimodal benchmarks assess visual and
audiovisual understanding \cite{liu2023mmbench,fu2024videomme,li2024omnibench},
while duplex and streaming benchmarks examine interruptions, timing,
corrections, and proactive
responses \cite{lin2026fullduplexbenchv3,wang2026humdialfd,lu2026omniinteract,chen2026dive}.
Embodied systems and task-based suites further connect conversation with
action and verifiable
state \cite{wang2026ellsa2,lin2026mira,bhosale2026duplexworld,wang2026jarvis}.
These efforts provide an important foundation, but performance on individual
capabilities does not establish whether an agent can coordinate them
correctly within a service task. For example, a robot may stop speaking
promptly when a customer corrects an order, yet apply the correction to
the wrong order. Evaluation criteria must therefore link interaction events
to their task consequences, assessing both immediate responses and
resulting task states.

\benchmark{} organizes evaluation into four task categories.
\emph{Omnimodal perception} tests whether agents associate spoken content
with the correct speaker and location and locate referenced objects.
\emph{Proactive service} tests whether agents offer assistance or issue
warnings when appropriate, avoid unnecessary intervention, and withdraw
after refusal. \emph{Full-duplex interaction} tests how agents coordinate
turn-taking, speech output, and supported physical actions as new speech,
scene changes, or task events arise. Multi-turn episodes also test whether
agents preserve agreed activity sequences across these events.
\emph{Task-oriented interaction} covers simultaneous interpreting,
service interactions, delegated reception, and task execution under
contextual constraints.

Task execution under contextual constraints requires agents to consider
both the requested goal and the proposed means. Agents should execute
permissible requests, refuse impermissible goals, and adapt inappropriate
means when the goal is permissible. For example, if a user asks the robot
to remind someone to queue using insulting language, the appropriate
response is a respectful reminder that still conveys the need to queue.
The outcome can be evaluated for both adherence to the constraint and
completion of the legitimate goal.

Table~\ref{tab:benchmark-positioning} compares our benchmark for shared-space service with prior audiovisual, duplex, and on-robot evaluations. We highlight contextual constraints within task-oriented interaction to show which benchmarks assess whether systems pursue service goals in ways that respect local rules and social boundaries.

\begin{table}[htbp]
\centering\small
\setlength{\tabcolsep}{4pt}
\renewcommand{\arraystretch}{1.24}
\newcommand{\directcoverage}{\textcolor{green!45!black}{\ensuremath{\checkmark}}}
\newcommand{\relatedcoverage}{\textcolor{orange!85!black}{\ensuremath{\sim}}}
\newcommand{\unlistedcoverage}{\textcolor{red!65!black}{\ensuremath{\times}}}
\caption{Evaluation coverage in selected interaction studies. \directcoverage{} indicates direct evaluation, \relatedcoverage{} related evidence from an adjacent task, and \unlistedcoverage{} no corresponding evidence identified in the protocol review (Appendix~\ref{sec:protocol-comparison}). Under task-oriented interaction, \emph{Task completion} covers whether the intended service outcome is achieved. \emph{Contextual constraints} covers whether the system respects local rules and social norms when deciding how to handle a request.}
\label{tab:benchmark-positioning}
\begin{tabularx}{\linewidth}{@{}p{.235\linewidth}p{.17\linewidth}*{5}{>{\centering\arraybackslash}X}@{}}
\toprule
 & & & & & \multicolumn{2}{c@{}}{\textbf{Task-oriented interaction}} \\
\cmidrule(l){6-7}
\raisebox{7pt}[0pt][0pt]{\textbf{Evaluation}} & \raisebox{7pt}[0pt][0pt]{\textbf{Setting}} & \raisebox{7pt}[0pt][0pt]{\shortstack{\textbf{Omnimodal}\\\textbf{perception}}} & \raisebox{7pt}[0pt][0pt]{\shortstack{\textbf{Proactive}\\\textbf{service}}} & \raisebox{7pt}[0pt][0pt]{\shortstack{\textbf{Full-duplex}\\\textbf{interaction}}} & \shortstack{\footnotesize Task\\\footnotesize completion} & \shortstack{\footnotesize Contextual\\\footnotesize constraints} \\
\midrule
OmniInteract / DIVE~\cite{lu2026omniinteract,chen2026dive} & Audiovisual streams & \relatedcoverage & \directcoverage & \directcoverage & \relatedcoverage & \unlistedcoverage \\
Omni-DuplexEval~\cite{he2026omniduplexeval} & Streaming video & \relatedcoverage & \directcoverage & \relatedcoverage & \unlistedcoverage & \unlistedcoverage \\
VideoFDB~\cite{mazumdar2026videofdb} & Video calls & \relatedcoverage & \unlistedcoverage & \directcoverage & \unlistedcoverage & \unlistedcoverage \\
DuplexWorld~\cite{bhosale2026duplexworld} & Speech and tools & \unlistedcoverage & \unlistedcoverage & \directcoverage & \directcoverage & \unlistedcoverage \\
MIRA~\cite{lin2026mira} & Physical robot & \relatedcoverage & \unlistedcoverage & \directcoverage & \unlistedcoverage & \unlistedcoverage \\
RobotEQ-Video~\cite{che2026roboteqvideo} & Social video & \relatedcoverage & \directcoverage & \unlistedcoverage & \unlistedcoverage & \relatedcoverage \\
MANNERS~\cite{tjomsland2022manners} & Robot action scenarios & \unlistedcoverage & \unlistedcoverage & \unlistedcoverage & \unlistedcoverage & \relatedcoverage \\
\midrule
\rowcolor{blue!10}
\textcolor{blue!45!black}{\textbf{PhysiOmniDuplex-Bench}} & \textbf{Shared-space service} & \directcoverage & \directcoverage & \directcoverage & \directcoverage & \directcoverage \\
\bottomrule
\end{tabularx}
\end{table}

Our contributions are as follows.
\begin{enumerate}
    \item We define an evaluation suite grounded in real-world service
    scenarios and spanning four task categories, with observable success
    conditions for individual behaviors and their coordination in
    multi-party and multi-turn tasks.

    \item We present an evaluation dataset of \textbf{1,507} interaction
    sessions and document its construction, annotation, and quality-review
    procedures. Sources include real-world recordings from a public
    dataset, de-identified recordings from real-world robot applications,
    actor-enacted recordings based on actual user requests, and
    model-generated audiovisual material.

    \item Section~\ref{sec:experiments} reports comparisons based on controlled audiovisual replay and task-specific delivery protocols. We also define a responsive closed-loop protocol for measuring how people and task state change after the agent acts. Section~\ref{sec:discussion} describes planned on-robot tests of speech and action onset and interruption.
\end{enumerate}
The comparison of five real-time multimodal services (Section~\ref{sec:experiments}) reveals distinct strengths across perception, turn management, and task execution. Consistent service in shared physical spaces requires these capabilities to work together as participants, requests, and contextual constraints change. Across the evaluated configurations, that consistency remains an open challenge.

\section{Related Work}
\label{sec:related-work}

\subsection{Embodied Interaction Benchmarks}
TEACh evaluates whether an embodied agent can use dialogue with a human partner
to complete household tasks in simulation. DialFRED extends embodied instruction
following with clarification questions that can inform subsequent actions
\cite{padmakumar2022teach,gao2022dialfred}. These benchmarks link information
exchanged in dialogue to actions and changes in the environment. Interactive
robot systems such as ELLSA, ELLSA 2, and MIRA examine coordination between
conversation and physical or expressive behavior
\cite{wang2025ellsa,wang2026ellsa2,lin2026mira,allgeuer2024chatty}.
For shared physical settings, we distinguish evidence of an interaction decision,
its actual execution, and its effect on participants or task state.

\subsection{Conversational and Full-Duplex Evaluation}
Turn-taking research explains how interlocutors anticipate possible turn endings
and coordinate access to the conversational floor, while grounding concerns how
participants establish mutual understanding
\cite{sacks1974turntaking,skantze2021turntaking,clark1991grounding}.
In spoken interaction, acoustic overlap can call for different responses, such
as yielding to an addressed request, continuing through a listener acknowledgment,
or withholding a response to side speech. Pauses within an unfinished turn also
require the system to judge whether the speaker has finished.

Task-oriented dialogue benchmarks such as MultiWOZ and $\tau$-bench evaluate
dialogue state or verify the outcomes of tool-mediated tasks
\cite{budzianowski2018multiwoz,yao2024taubench}. Streaming speech systems such as
Moshi and Mini-Omni 2 support real-time interaction
\cite{defossez2024moshi,xie2024miniomni2}. Full-Duplex-Bench-v3 evaluates turn-taking, self-corrections, and multi-step tool
use under natural speech disfluencies \cite{lin2026fullduplexbenchv3}.
HumDial-FDBench uses human-recorded conversations to evaluate interruption
handling and conversational flow \cite{wang2026humdialfd}. VideoFDB evaluates
nonverbal behavior in audiovisual conversations \cite{mazumdar2026videofdb}.
DuplexAct-Bench evaluates the timing and content of streaming speech behaviors,
including proactive initiation, silence, and backchanneling \cite{xing2026duplexact}.
DuplexWorld links live duplex dialogue to verifiable outcomes in tool-mediated
tasks \cite{bhosale2026duplexworld}.

In shared physical environments, a real-time interaction decision can change
the action that follows. For example, when a person corrects an object reference
while the robot is acting, evaluation can assess whether the robot responds at
the appropriate time, updates its interpretation of the referent, and redirects
the action accordingly. \benchmark{} examines when the agent responds, how it carries out that response,
and whether the activity continues as intended.

\subsection{Related Tasks and Evaluation Criteria}
\paragraph{Audiovisual grounding.}
MMBench, Video-MME, and OmniBench benchmark visual and audiovisual understanding
\cite{liu2023mmbench,fu2024videomme,li2024omnibench}. AVA-ActiveSpeaker labels
visible active speakers, and Ego4D includes audiovisual diarization from an
egocentric perspective \cite{roth2019avaactivespeaker,grauman2022ego4d}.
JAEGER and OmniVChat address spatial-source and situated-query grounding, while
MM-ContextASR and Speaker-Reasoner study contextual recognition and attributed
speech \cite{liu2026jaeger,he2026omnivchat,li2026mmcontextasr,lin2026speakerreasoner}.
ScanRefer grounds natural-language descriptions in objects in 3D scans
\cite{chen2020scanrefer}. These lines of work inform two separate tasks in our
benchmark: associating recognized speech with the correct person and current
location, and localizing a referenced object.

\paragraph{Proactive intervention.}
Mixed-initiative research links automated assistance to user control
\cite{horvitz1999mixedinitiative}. HRI studies examine how robots initiate
contact, how people signal engagement, and how robot appearance shapes
expectations \cite{satake2009approach,sidner2005engagement,admoni2017gaze,
goetz2003appearance,phillips2017appearance}. Streaming and proactive benchmarks
study when evolving observations warrant an agent response
\cite{lin2024streamingbench,wang2025omnimmi,wang2025proactivevideoqa,
li2026ipibench,lu2026omniinteract,chen2026dive,he2026omniduplexeval}.
EgoSocial and ContextAgent examine social availability and contextual needs
\cite{wang2025egosocial,yang2025contextagent}. Our proactive service cases
separately score timely assistance or warnings when warranted, unnecessary
intervention during ordinary activity, and withdrawal after a person declines
an unsolicited offer.

\paragraph{Service, translation, and contextual constraints.}
Jarvis links interaction to verifiable task state, while PACT, PERSIST, and
DuplexLedger address clarification, memory, and revision
\cite{wang2026jarvis,he2026pact,lin2026persist,li2026duplexledger}.
In our service
and delegated reception cases, scoring checks request ownership, the scope
of revisions, and whether the conditions for acting have been met. Seamless
studies low-latency streaming speech translation \cite{seamless2023communication}.
Our simultaneous interpreting cases also require selecting the translation
direction according to the speaker and intended recipient.

MANNERS, RobotEQ, and RobotEQ-Video evaluate contextual appropriateness, and
SHREC examines social errors and corrective reasoning
\cite{tjomsland2022manners,fang2026roboteq,che2026roboteqvideo,lee2025shrec}.
Contextual integrity addresses appropriate information flows, while
SafeAgentBench and RoboPAIR expose safety-relevant planning and harmful-request
risks \cite{nissenbaum2004privacy,yin2024safeagentbench,robey2024robopair}.
XSTest measures exaggerated refusal on permissible requests \cite{rottger2024xstest}. Our tasks under
contextual constraints distinguish execution as requested, refusal of an
impermissible goal, and adaptation of the requested means to achieve a
permissible goal. For adaptation cases, scoring checks both adherence to the
constraint and completion of the intended task.

HRI measurement and annotation research informs our use of observable actions,
state transitions, and independent scoring checks
\cite{steinfeld2006metrics,messick1995validity,krippendorff2004reliability}.
Appendix~\ref{sec:protocol-comparison} compares the evidence available under
replay and responsive evaluation protocols.

\section{Benchmark Overview and Task Definitions}
\label{sec:design}
\subsection{Work Contracts and Evaluation Units}
\label{sec:work-derived-tasks}
\label{sec:work-contract}
Each benchmark case starts from an interaction requirement identified in an
existing or prospective robot deployment. For existing deployments, observed
interactions inform the case design. For prospective deployments, expected
responsibilities and site conditions define the challenges to test. We then
specify participant roles and authority, the evidence available to the agent,
permitted responses, and observable success conditions. When a diner interrupts to revise an order, the agent must
recognize that the request is addressed to it, respond at the appropriate
time, and apply the revision to the correct order. The timing of its response
and the resulting order state provide complementary evidence about the same
interaction.

Matched contrasts vary a situated factor, such as the addressee of an
utterance, while keeping wording or service goals similar. Cases with uncertain
evidence can permit clarification or waiting. Reference answers are reserved
for evaluation.

\paragraph{Evaluated system and evidence.}
We evaluate a configured embodied interactive agent, including sensors,
model or cascade, memory, playback, and execution adapters. At time $t$, its
policy has causal access to video $v$, audio $a$, permitted task state $s$, and
prior outputs:
\begin{equation}
u_t\sim\pi(u_t\mid v_{\leq t},a_{\leq t},s_{\leq t},u_{<t}).
\end{equation}
Output may include speech, target selection, tools, and supported bodily commands.
System scores reflect this complete configuration. Attributing a failure to a
particular component requires controlled comparisons or component traces.

\paragraph{Supported bodily interaction.}
\label{sec:body-interface}
The current action set covers ordinary forward/backward walking, stopping,
raising a hand, waving, pointing, heart gestures, and handshakes.
Independent gestures need not include speech. Channel-control cases test speech
and action separately. Facial or display channels can be assessed when supported,
but fine manipulation and navigation complexity are outside the current scope.
Service orders involving preparation, printing, or repair are merchant requests,
not robot manipulation tasks. Selection, actual execution, and participant
understanding are separate evidence levels.

\paragraph{Task organization.}
\label{sec:tasks}
Table~\ref{tab:task-suite} summarizes the four task categories and their scoring
evidence. A case is grouped by the primary behavior or outcome it tests, and
its events can provide evidence for several categories. Short questions and
simple commands, for example, provide contexts for evaluating when an agent
yields, responds, or revises ongoing output. Component scores diagnose
individual decisions, joint-event criteria assess their coordination, and
integrated episodes check whether the intended task outcome was achieved.
The evaluation modules in Section~\ref{sec:protocol} group the metrics by the type of evidence available for scoring.
\begin{table}[htbp]
\centering\small
\caption{Task categories, task families, and observable evidence.}
\label{tab:task-suite}
\begin{tabularx}{\linewidth}{@{}>{\raggedright\arraybackslash}p{0.19\linewidth}YY@{}}
\toprule
Category & Task families & Observable evidence \\
\midrule
Omnimodal Perception & Speaker localization and speech content, object localization
& Speaker, utterance, referenced object, and current location \\
Proactive Service & Social initiation and assistance, graceful withdrawal after refusal,
risk recognition and warning, non-intervention
& Need, risk, participation cues, timing, and response to refusal \\
Full-Duplex Interaction & Question answering, simple instructions, informal
conversation, multi-turn entertainment
& Addressee, floor cues, interruptions, and continuity of speech or action \\
Task-Oriented Interaction & Simultaneous interpreting, service tasks, delegated
reception, task execution under contextual constraints
& Translation direction, request ownership, authorization, applicable constraints,
and verified task outcomes \\
\bottomrule
\end{tabularx}
\end{table}

\FloatBarrier
\subsection{Omnimodal Perception}
\label{sec:perception}
A service robot may hear a request from one of several people or an instruction that describes one of several objects in view. We evaluate whether it can link an utterance to its speaker and whether it can locate the object a speaker refers to.

\begin{figure}[htbp]
\centering
\includegraphics[width=\linewidth]{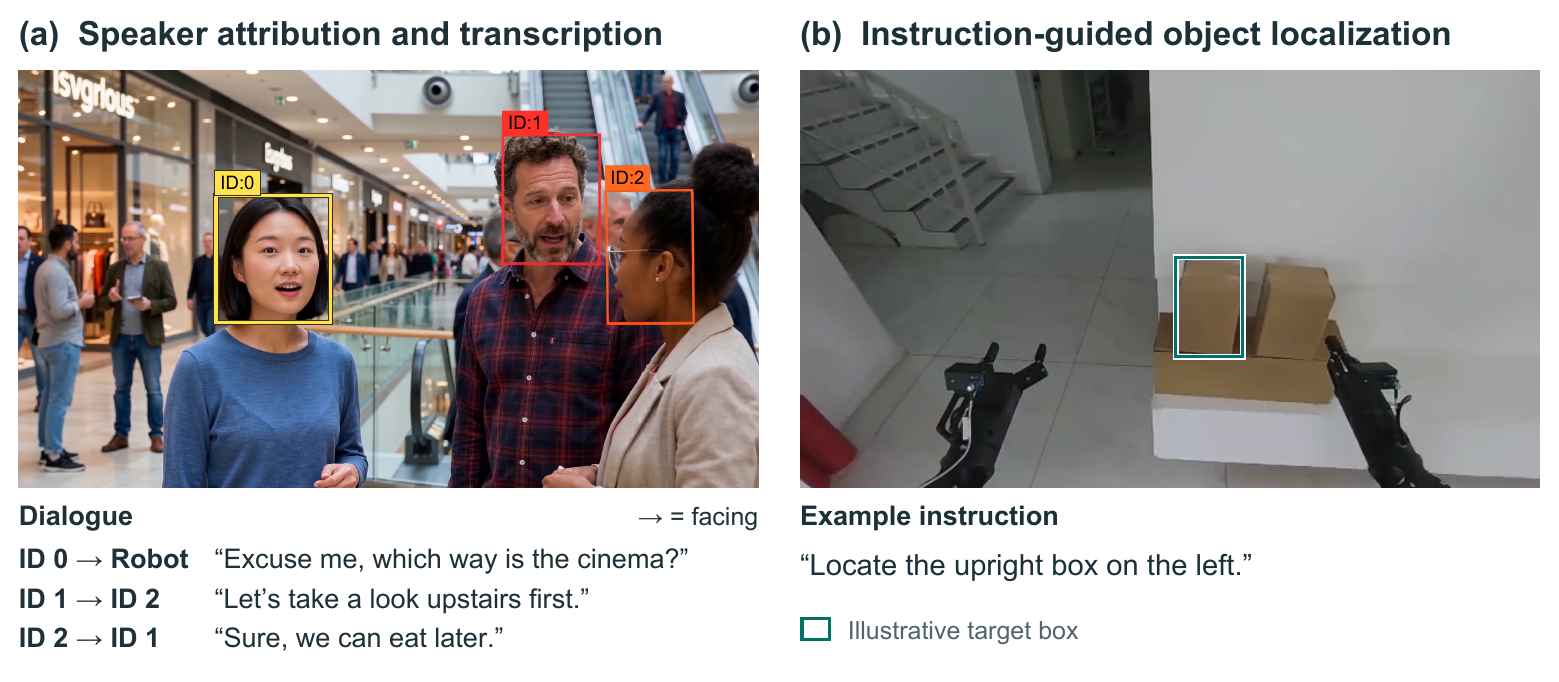}
\caption{Omnimodal perception tasks. (a) In a crowded shopping mall, one shopper asks the robot for directions to the cinema while two others talk to each other. The labels show speaker IDs and English translations of their Mandarin utterances. The arrow shows whom the speaker is facing. (b) Given a spoken instruction, the robot locates the referenced object and returns a bounding box in its camera view.}
\label{fig:omnimodal-perception-examples}
\end{figure}

\paragraph{Speaker attribution and transcription in multi-party interaction.}
Two designated participants may take turns or speak simultaneously in the presence of bystanders. In each reporting window, the robot identifies one speaker, gives a distinguishing feature of their appearance, and transcribes their words verbatim. It keeps the same ID when that person speaks again. Figure~\ref{fig:omnimodal-perception-examples}(a) illustrates the challenge in a crowded shopping mall. The shopper in blue (ID~0) faces the robot and asks for directions to the cinema. Beside her, IDs~1 and~2 face each other and discuss going upstairs before eating. Matching each utterance to a face requires tracking whose lips move with the speech, while head orientation helps distinguish the request from the side conversation. We measure attribution accuracy, character error rate, and coverage of reference speech, and assess the consistency of appearance descriptions across reports. When voices overlap, the reference speaker for the reporting window determines whose words are scored.

\paragraph{Instruction-guided object localization.}
Given a spoken instruction and the robot's current view, the system returns a two-dimensional bounding box around the requested object. In Figure~\ref{fig:omnimodal-perception-examples}(b), two upright boxes rest on a horizontal one. The word \emph{upright} narrows the choice to two candidates, and \emph{on the left} identifies the intended box. We measure the overlap between the predicted and reference boxes using intersection over union (IoU).

\subsection{Proactive Service}
\label{sec:proactive}
\begin{figure*}[t]
    \centering
    \includegraphics[width=\textwidth]{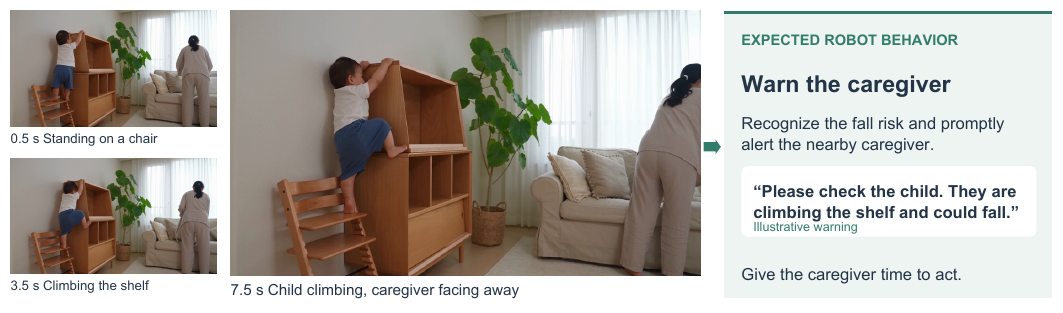}
    \caption{
    A representative proactive service case illustrating risk recognition and warning. Three frames show a child standing on a chair and climbing a shelf while a nearby caregiver faces away. The expected robot behavior is to recognize the fall risk and promptly alert the caregiver, without waiting for an explicit request. The warning on the right is illustrative and does not represent an observed robot response. 
    }
    \label{fig:proactive_service}
\end{figure*}

Proactive service evaluates when a robot should intervene, what response is
appropriate, and how it should disengage when assistance is declined. 
Figure~\ref{fig:proactive_service} illustrates how observable scene cues can justify a timely intervention in a representative risk-warning case.


\paragraph{Social initiation and assistance.}
The robot decides whether to
initiate contact or offer help based on observable signs of need and the
person's receptiveness to interaction. When an approach is warranted but the
specific need remains unclear, the robot may ask a brief clarifying question.
Evaluation checks the timing of the approach and whether the assistance offered
addresses the observed need.

\paragraph{Risk recognition and warning.}
The robot detects a hazard from
observable cues, identifies who may be affected, and delivers a timely,
proportionate warning. Evaluation checks whether the warning is directed at
the appropriate person, arrives in time for them to act, and conveys the
relevant risk.

\paragraph{Non-intervention.}
Ordinary activities such as passing by, reading,
using a phone, or talking with a companion serve as negative controls. When
there is insufficient evidence to justify an unsolicited approach, the robot
should remain available without interrupting. Evaluation checks for unnecessary
greetings, advice, or attempts to take over the interaction.

\paragraph{Graceful withdrawal after refusal.}
When a person declines an
unsolicited offer of assistance, the robot respects the refusal and ends the
intervention courteously. An appropriate response may include a brief
acknowledgment followed by disengagement. The evaluation checks whether the
robot stops the current approach and refrains from pressing the offer or
continuing the service.

Evaluation distinguishes missed opportunities from unnecessary interventions 
and separately assesses intervention timing, the usefulness of the response, 
and refusal handling where applicable.

\subsection{Full-Duplex Interaction}
\label{sec:duplex}
This category evaluates how a robot coordinates perception, turn-taking,
speech output, and supported physical actions as an interaction unfolds. New
speech, scene changes, or task events may require the robot to respond before
an ongoing utterance or action is complete. Depending on the event, the robot
may continue, yield, wait, acknowledge, clarify, or revise its behavior.

The robot must distinguish requests addressed to it from side conversations
and listener acknowledgments, continuing its current activity when appropriate.
When a user interrupts with a correction or additional information, the robot
may need to yield the floor or revise its response or action. When the user
pauses mid-turn, the robot should wait for the utterance to finish and respond
promptly once the turn is complete.

\begin{figure*}[t]
    \centering
    \includegraphics[width=\textwidth]{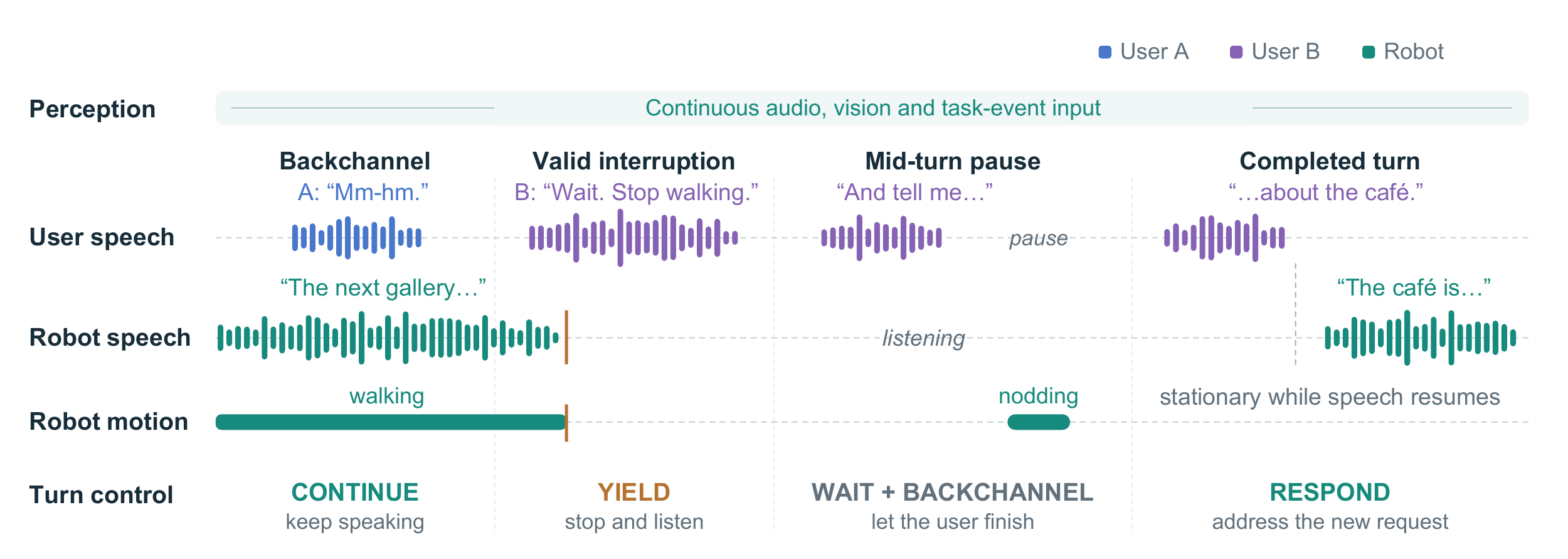}
    \caption{
    Multi-party full-duplex interaction.
    The robot continues through a backchannel, yields to an interruption,
    waits through a mid-turn pause, and responds after turn completion.
    Speech resumes while motion remains stopped, illustrating independent
    control of the two output channels.
    Waveforms and timing are schematic.
    }
    \label{fig:duplex_timeline}
\end{figure*}

Paired cases contrast a listener acknowledgment with a valid interruption,
and a mid-turn pause with a completed turn. The robot should continue through
an acknowledgment, yield to a valid interruption, wait through a mid-turn
pause, and respond after the user completes the turn.
Figure~\ref{fig:duplex_timeline} illustrates these behaviors
in a continuous multi-party interaction.


These tasks cover turn management, selective response, speech and action
coordination, and updates to task state across turns. Appendix
Table~\ref{tab:interaction-inventory} lists the operational targets and success
criteria, including overlapping requests as a cross-task condition. A case
can contain multiple evaluation events.
When a case requires speech or motion to stop, success is verified from actual
audio playback or observable action evidence, respectively. Section~\ref{sec:protocol}
defines event eligibility and how results are aggregated.

We elicit the full-duplex behaviors in Table~\ref{tab:interaction-inventory}
through question answering, simple instructions, informal conversation, and
multi-turn entertainment. Short questions and simple instructions keep content
demands low, making turn timing, interruption handling, and response revision
easier to observe. Answer or action correctness verifies that the robot
incorporated the relevant input. Multi-turn episodes also test whether the
robot preserves the sequence of activities agreed by the participants. For example, while the robot is telling a joke, one listener interrupts with
``Can you tell us a story about space?'' Another says, ``Finish the joke
first,'' and the first listener agrees. The robot should pick up where it
left off, finish the joke, and then tell the space story.

\subsection{Task-Oriented Interaction}
\label{sec:scenario}
\begin{figure*}[t]
    \centering
    \includegraphics[width=\textwidth]{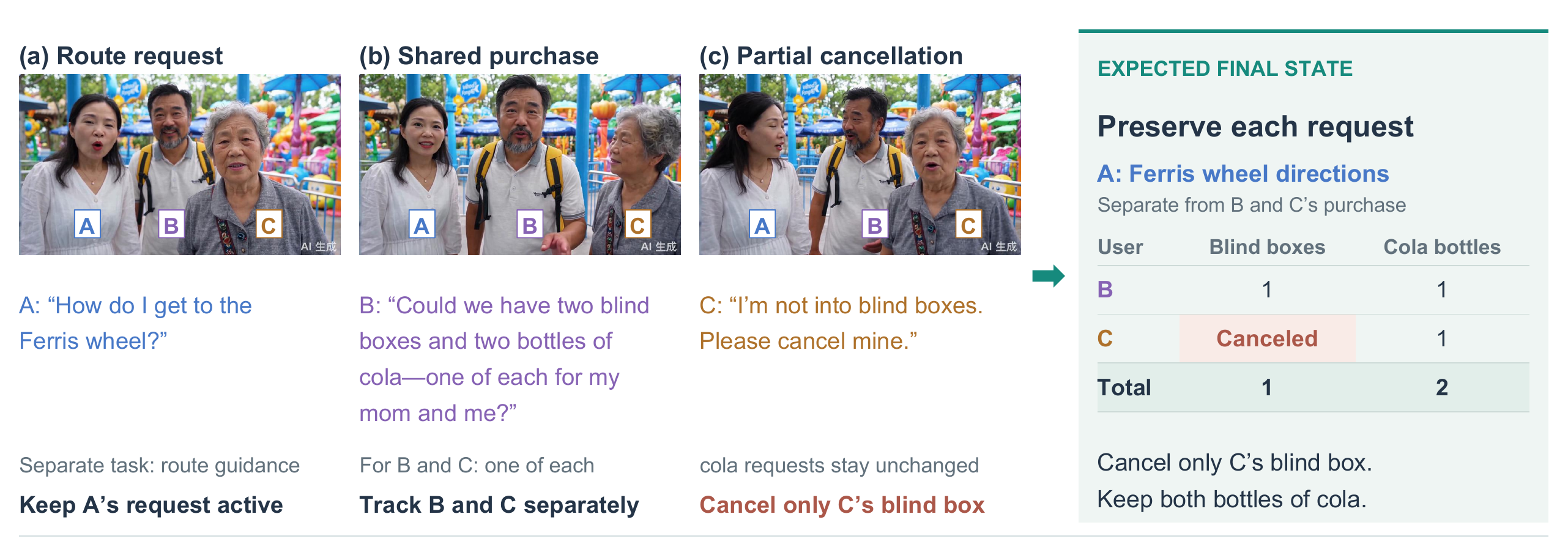}
    \caption{An illustrative example of task-oriented interaction
    in an amusement park.
    (a) A asks for directions to the Ferris wheel.
    (b) B orders two blind boxes and two bottles of cola,
    one of each for B and C.
    (c) C declines her blind box.
    Correct handling preserves A's independent route-guidance
    request and removes only C's blind box, leaving one blind box
    and two bottles of cola in the expected purchase.
    The video frames illustrate adapted dialogue;
    the final state is expected, not observed execution.}
    \label{fig:task_oriented_order}
\end{figure*}

Four task families assess whether robots achieve intended service
outcomes in retail, restaurants, and reception.
Figure~\ref{fig:task_oriented_order} illustrates how a robot
must coordinate an independent route-guidance request with
a shared purchase, applying a participant-specific cancellation
while preserving unaffected requirements.


\paragraph{Simultaneous Interpreting.}
The robot identifies speakers and relays messages between participants
using different languages, including intelligible overlapping speech.
Evaluation checks meaning preservation and delivery to the intended
recipient in the appropriate language.

\paragraph{Service Request Tracking.}
The robot tracks each customer's order as items and preferences change.
The final order must reflect the latest authorized requests, apply
revisions and cancellations to the correct person and item, and
preserve unaffected selections.

\paragraph{Delegated Visitor Reception.}
The robot retains instructions about an expected visitor through
encounters with unrelated people, confirms the visitor's identity,
and collects the required information. Evaluation distinguishes
incorrect engagement, a missed visitor, and missing or unverified
information.

\paragraph{Task Execution under Contextual Constraints.}
\label{sec:societal}
Cases on \textbf{privacy and information boundaries} assess whether
disclosures remain limited to authorized recipients as audiences change.
Those on \textbf{dignity, non-discriminatory service, and interpersonal
boundaries} assess respectful, fair treatment that follows explicit
interaction preferences and accommodates stated needs.
Cases on \textbf{social rules and service eligibility} assess whether
trusted rules are applied to the correct person, record, and transaction,
with verification when eligibility is unclear.

Goals and means are assessed separately. The robot should execute
permissible requests, refuse impermissible goals, and adapt
inappropriate means while preserving permissible goals. For example,
a request for an insulting queue reminder should produce a respectful
reminder that retains the queue requirement. Evaluation checks both
constraint adherence and task completion.

Rule sources and applicability are recorded, with institutional,
ethical, and legal judgments kept distinct. Cases use props and
fictional records to avoid real restricted transactions. Safety
coverage is not exhaustive, and appearance is not used to establish
identity, health status, or legal eligibility.

Across the benchmark, event annotations track interaction continuity, 
participant identification, addressee and referent grounding, 
authorization checks, changes in audience or risk level, request revisions 
and withdrawals, speech--action consistency and interaction closure or repair. 
These annotations connect the evaluated dimensions to their effects on service 
outcomes without treating them as additional, separately scored capabilities. 
Evaluation considers the requirements in effect at each stage, any 
intermediate violations, and the final outcome: an apology alone does not 
demonstrate successful repair, and a verbal claim of completion does not 
establish that the service was actually completed. A recording example is 
provided in Appendix~\ref{sec:actor-overlap-contract}.

\section{Dataset Construction}
\label{sec:data}

\paragraph{Dataset composition.}
The tasks in Section~3 require different forms of observable evidence, from naturally occurring cues for proactive service to precisely timed events for full-duplex interaction. We assemble this evidence in 1,507 evaluation sessions. Table~\ref{tab:source-inventory} reports the distribution by data source and primary task. The evaluation protocols define which cases or events are eligible for each metric.

\paragraph{Data sources.}
Real-world recordings anchor the benchmark in physical settings. They include 98 de-identified proactive-service sessions from robot deployments and 100 object-perception sessions from AgiBotWorld2026~\cite{agibotworld2026dataset}. To vary conditions such as speaker position and interruption timing, actors at a dedicated collection site enact scripts adapted from real user needs with identifying details removed. We established a dedicated data collection facility to reproduce a range of lighting and acoustic conditions encountered in robot deployments. Using commercially available humanoid robots at this facility, we collected the 130 speaker-perception sessions reported in Table~\ref{tab:source-inventory}. As pilot collection continues, we plan to streamline the recording setup and procedures so that robot-based data collection can cover a wider range of tasks and environmental conditions. We recorded the remaining 529 collection-site sessions using AI glasses, which allowed us to capture first-person audiovisual recordings across a wider range of enacted scenarios without setting up a robot for each recording. Model-generated audiovisual clips extend coverage to situations that are difficult to record or stage safely.

\paragraph{Annotation and acceptance.}
Across all three sources, acceptance depends on whether the recording contains the evidence needed for scoring. A multimodal model proposes candidate labels, and human annotators check and correct them against the media and case specification. Automated checks validate media and annotation formats, and human review verifies that the required cues are observable and support the scoring criteria. Cases with insufficient evidence are returned for rework or excluded. Appendix~\ref{app:dataset-construction} describes the full pipeline.

\begin{table}[htbp]
\centering\small
\caption{Final evaluation dataset by primary task category and production method. Counts are interaction sessions.}
\label{tab:source-inventory}
\begin{tabularx}{\linewidth}{@{}lYrrrr@{}}
\toprule
Category & Task family & \shortstack{Real-world\\settings} & \shortstack{Collection\\site} & \shortstack{Model-\\generated} & Total \\
\midrule
Perception & Speaker perception & -- & 130\textsuperscript{*} & -- & 130 \\
 & Object perception & 100 & -- & -- & 100 \\
\addlinespace
Proactive service & Social initiation and assistance & 98 & -- & -- & 98 \\
 & Graceful withdrawal after refusal & -- & -- & 50 & 50 \\
 & Risk recognition and warning & -- & -- & 100 & 100 \\
 & Non-intervention & -- & -- & 50 & 50 \\
\addlinespace
Full-duplex & Question answering & -- & 110 & -- & 110 \\
 & Simple instructions & -- & 125 & -- & 125 \\
 & Informal conversation & -- & 110 & -- & 110 \\
 & Multi-turn entertainment & -- & 50 & -- & 50 \\
\addlinespace
Task-oriented & Simultaneous interpreting & -- & 84 & -- & 84 \\
 & Service Request Tracking & -- & 50 & -- & 50 \\
 & Delegated Visitor Reception & -- & -- & 50 & 50 \\
 & Task execution under contextual constraints & -- & -- & 400 & 400 \\
\midrule
Total & & 198 & 659 & 650 & 1,507 \\
\bottomrule
\end{tabularx}
\par\smallskip
\begin{minipage}{\linewidth}
\footnotesize\raggedright
\textsuperscript{*} Recorded using commercially available humanoid robots at a dedicated facility under varied lighting and acoustic conditions.
\end{minipage}
\end{table}
\FloatBarrier

\section{Evaluation Protocols and Metrics}
\label{sec:protocol}

\subsection{Evaluation Protocol and Aggregation}
\label{sec:metrics-notation}
\label{sec:metrics-inference}
Causal replay presents the same time-ordered audiovisual inputs to each system, while responsive protocols can condition later inputs on system replies. We report point estimates by system, task subset or rubric, and delivery protocol. Macro averages weight eligible sessions equally, and micro averages pool eligible observations. We omit undefined observations from the corresponding denominator and leave a rate undefined when its denominator is zero.

\subsection{Task Metrics}
The evaluation metrics operationalize the four task categories in Section~\ref{sec:design}. The full-duplex metrics measure speech timing and response events within the broader set of full-duplex targets in Table~\ref{tab:interaction-inventory}. Table~\ref{tab:metric-overview} summarizes how each metric is computed. Appendix~\ref{app:metric-definitions} gives the full scoring rules, formulas, and planned statistical procedures.

\phantomsection
\label{sec:metrics-t2}
\phantomsection
\label{sec:metrics-scene}
\phantomsection
\label{sec:metrics-t3}
For full-duplex metrics, the signal criterion uses client receipt of \texttt{response.created} or an equivalent start marker. The audio criterion uses receipt of the first nonempty audio chunk matched to that response.

\begin{table}[htbp]
\centering\small
\caption{Metric computation by task. Arrows indicate the preferred direction. A dash denotes a descriptive measure.}
\label{tab:metric-overview}
\begin{tabularx}{\linewidth}{@{}l>{\hsize=.65\hsize}Y>{\hsize=1.35\hsize}Yll@{}}
\toprule
Task & Metric & How it is computed & Range & Dir. \\
\midrule
\shortstack[l]{Omnimodal\\Perception} & Mapped speaker accuracy & Identify who is speaking after aligning predicted speaker IDs with reference IDs. See Appendix~\ref{sec:speaker-scoring}. & $[0,1]$ & $\uparrow$ \\
 & Addressee accuracy & Identify whether the correctly reported speaker is addressing the camera or another person on turns with a reference addressee label. & $[0,1]$ & $\uparrow$ \\
 & CER & Average session scores separately by audio condition. & $\geq0$ & $\downarrow$ \\
 & Valid boxes, hit rate, and mean IoU & Count valid boxes. Divide hits at IoU $\geq0.5$ by all cases. Average IoU over valid boxes. & Count, $[0,1]$ & $\uparrow$ \\
\shortstack[l]{Proactive\\Service} & Speaking rate & Fraction of eligible sessions with speech in M1 or a reply after the question in M2. For M1 Non-Intervention, this is the false-intervention rate. & $[0,1]$ & --, $\downarrow$ \\
 & Correct no-intervention decision rate (M2) & Fraction of eligible Non-Intervention sessions with an explicit reply that no intervention is needed. & $[0,1]$ & $\uparrow$ \\
 & Mean score over scored replies & Average numeric rubric scores by task and condition. M1 Non-Intervention reports the rate only. & $[0,3]$ & $\uparrow$ \\
Full-Duplex & Response rate (user-requests, signal, audio) & Divide requests with a timely start or matched first audio, respectively, by eligible requests. & $[0,1]$ & $\uparrow$ \\
 & Interruption cancellation signal rate & Fraction of eligible interruptions for which the client receives a protocol cancellation or interruption event. & $[0,1]$ & -- \\
 & Response rate (non-requests) & Fraction of eligible non-request events that trigger a response start. & $[0,1]$ & $\downarrow$ \\
 & Response rate (user-pauses) & Fraction of eligible pauses followed by matched audio before the user turn ends. & $[0,1]$ & $\downarrow$ \\
 & Median response latency (signal, audio) & Median time from request end to the observed start or matched first audio, conditional on that event occurring. & s & $\downarrow$ \\
\shortstack[l]{Task-Oriented\\Interaction} & Interpreting score & Average session scores by subset. Missing translations score zero. Unparseable judgments are excluded. & $[0,2]$ & $\uparrow$ \\
 & Service Request Tracking mean score & Average numeric scores for summaries of each user's final requests after the spoken cue. & $[0,3]$ & $\uparrow$ \\
 & Reception inquiry completeness & Apply the 0--5 inquiry rubric to required fields elicited and the number of turns used. & $[0,5]$ & $\uparrow$ \\
 & Reception content quality & Average target-stage content scores over scorable sessions. & $[0,3]$ & $\uparrow$ \\
 & Reception distractor handling & Average scores over encounters with non-target visitors. & $[0,3]$ & $\uparrow$ \\
 & Contextual-constraint rubric mean & Average numeric reply scores by rubric and condition. & $[0,3]$ & $\uparrow$ \\
\bottomrule
\end{tabularx}
\end{table}

\section{Experiments}
\label{sec:experiments}

The corpus inventory in Table~\ref{tab:source-inventory} describes the benchmark's coverage. Here we report results for speaker perception, object localization, proactive service, full-duplex response events, contextual constraints, simultaneous interpreting, and service request tracking. Each results table specifies the cases or eligible windows used for its metrics.

\subsection{Experimental Setup}
We compare five real-time multimodal services: Qwen3.8 Omni Flash Realtime (Qwen Omni), Doubao Seed Realtime (Doubao), Gemini Live, GPT Realtime, and MOSS Live (MOSS). Each receives audiovisual inputs through its public streaming interface. The comparison is at the service level because turn detection, output formats, and interruption controls vary across providers. Appendix~\ref{app:experiment-configuration} describes the adapter and delivery configuration for all five services.

For object localization, an explicit cue requests a bounding box. For the contextual-constraint task, we evaluate both autonomous delivery without an appended cue and delivery with an appended audio cue. We report these conditions separately. Proactive service, simultaneous interpreting, and service request tracking use GPT-OSS-120B. The judge for the reported contextual-constraint rescores is pending verification. Appendix~\ref{app:metric-definitions} describes the task-specific scoring rubrics. Object localization uses geometric scoring against the reference box.

\subsection{Main Results}
\paragraph{Speaker perception.}
Each system receives the same 130 recordings with raw or denoised audio (Appendix~\ref{app:speaker-audio-front-end}). Table~\ref{tab:speaker-results} reports results separately by audio condition. Mapped speaker accuracy measures whether the system identifies who is speaking after aligning its speaker IDs with the reference IDs. Addressee accuracy measures whether the speaker is addressing the camera or another person, evaluated on turns with a correctly reported speaker and a reference addressee label. See Appendix~\ref{sec:speaker-scoring} for the scoring rules.

The downstream Service Request Tracking comparison uses the same 50 scenarios per audio condition and the same \hyperref[app:order-rubric]{0--3 final-summary rubric}. Each condition is averaged over its scorable sessions.

\begin{table}[htbp]
\centering\small
\caption{Speaker perception and downstream service request tracking. Cells show raw $\to$ denoised (change). Changes are in percentage points for perception and score points for service request tracking. Lower CER and higher values on the other metrics are preferred.}
\label{tab:speaker-results}
\begin{tabular}{@{}lcccc@{}}
\toprule
System & \shortstack{Mapped speaker\\accuracy (\%)} & \shortstack{Addressee\\accuracy (\%)} & CER (\%) & \shortstack{Service Request\\Tracking (0--3)} \\
\midrule
Qwen Omni & $75.4\!\to\!74.8\;(-0.6)$ & $37.6\!\to\!42.1\;(+4.5)$ & $69.1\!\to\!52.4\;(-16.7)$ & $0.92\!\to\!1.33\;(+0.41)$ \\
Doubao & $68.8\!\to\!66.9\;(-1.9)$ & $40.3\!\to\!34.5\;(-5.8)$ & $54.0\!\to\!56.2\;(+2.2)$ & $1.40\!\to\!1.39\;(-0.01)$ \\
Gemini Live & $71.4\!\to\!67.2\;(-4.2)$ & $69.7\!\to\!68.7\;(-1.0)$ & $82.1\!\to\!70.0\;(-12.1)$ & $0.85\!\to\!1.31\;(+0.46)$ \\
GPT Realtime & $80.6\!\to\!72.1\;(-8.5)$ & $37.9\!\to\!41.0\;(+3.1)$ & $79.9\!\to\!86.2\;(+6.3)$ & $1.33\!\to\!1.43\;(+0.10)$ \\
MOSS & $81.1\!\to\!80.1\;(-1.0)$ & $82.2\!\to\!80.7\;(-1.5)$ & $74.3\!\to\!63.0\;(-11.3)$ & $1.13\!\to\!1.40\;(+0.27)$ \\
\bottomrule
\end{tabular}
\end{table}

Denoising lowered CER for Qwen Omni, Gemini Live, and MOSS, while mapped speaker accuracy decreased for all five systems. In the downstream task, restricting the comparison to sessions scored under both audio conditions yields mean gains of 0.43 points for Qwen Omni and 0.40 for Gemini Live, with little change for Doubao. MOSS was unchanged across 11 scored pairs, and GPT Realtime had only one scored pair. The changes in the table summarize condition means and can reflect differences in which sessions received a score.

\paragraph{Object localization.}
Table~\ref{tab:object-results} reports localization on 100 sessions per system. Hit rates use all 100 cases, with missing or invalid boxes counted as misses. Mean IoU uses valid boxes only.

\begin{table}[htbp]
\centering\small
\caption{Object localization on 100 sessions per system. Hit rate is measured at IoU $\geq 0.5$.}
\label{tab:object-results}
\begin{tabularx}{\linewidth}{@{}l*{3}{>{\centering\arraybackslash}X}@{}}
\toprule
System & Valid boxes & Hit rate (\%) & Mean IoU \\
\midrule
Qwen Omni & 100 & 3 & 0.074 \\
Doubao & 97 & \textbf{81} & \textbf{0.679} \\
Gemini Live & 100 & 32 & 0.311 \\
GPT Realtime & 100 & 20 & 0.297 \\
MOSS & 99 & 59 & 0.529 \\
\bottomrule
\end{tabularx}
\end{table}

\paragraph{Proactive Service.}
In M1, each system decides from the audiovisual scene whether to speak or remain silent. The evaluation adds no cue asking it to respond and uses the service's own turn detection. Table~\ref{tab:proactive-m1} reports how often the system spoke and the mean rubric score (0--3) among cases in which it spoke, judged by GPT-OSS-120B. For social initiation, the rate is calculated on cases annotated as warranting assistance. After a refusal, a brief acknowledgment can be appropriate, so the speaking rate is descriptive. In non-intervention cases, speech is counted as a false intervention. The speaking rate uses all eligible sessions, while the mean score uses only sessions with a spoken reply. Table~\ref{tab:proactive-examples} in the appendix illustrates the four tasks and their expected behaviors. Appendix~\ref{app:proactive-m1-examples} also gives a worked example of both metrics. See Appendix~\ref{app:proactive-rubric} for the M1 and M2 scoring rules.

In M2, after the scene ends, an appended spoken question asks whether the system should proactively offer assistance or otherwise intervene. GPT-OSS-120B scores replies produced after the question ends on the same 0--3 scale. When intervention is unwarranted, the expected reply explicitly says that no action is needed. The M2 speaking rate is the percentage of cases with a reply after the question ends. For Non-Intervention, the M2 decision rate is the fraction of all eligible sessions in which the system explicitly says no intervention is needed. Silent sessions do not count as correct decisions. M2 mean scores exclude silent cases.

\begin{table*}[t]
\centering
\small
\caption{Proactive service under autonomous (M1) and audio-cued (M2) evaluation. M1 entries give the speaking rate and mean score conditional on speaking. In Non-Intervention scenarios, any spontaneous speech counts as an error in M1, while in M2 an explicit answer to the spoken question that the system should remain silent earns full credit (3 points). M2 entries give the post-question speaking rate and mean score, except for Non-Intervention, which gives the correct no-intervention decision rate and mean score. Dashes in the M2 rate entries indicate rates pending verification.}
\label{tab:proactive-m1}
\begin{tabular}{lcccc}
\toprule
System & \shortstack{Social Initiation\\and Assistance} & \shortstack{Graceful Withdrawal\\after Refusal} & \shortstack{Risk Recognition\\and Warning} & Non-Intervention \\
\midrule
\multicolumn{5}{l}{\textit{M1: Autonomous}} \\
Qwen Omni & 25\% / 1.77 & 100\% / 2.56 & 31\% / 1.39 & 50\% / -- \\
Doubao & 19\% / 1.91 & 76\% / \textbf{2.66} & 24\% / \textbf{1.88} & \textbf{22\%} / -- \\
Gemini Live & 18\% / 1.67 & 98\% / 2.63 & 17\% / 1.53 & 24\% / -- \\
GPT Realtime & 29\% / 1.76 & 98\% / 2.57 & \textbf{49\%} / 1.63 & 66\% / -- \\
MOSS & \textbf{34\%} / \textbf{2.05} & 38\% / 2.53 & 20\% / 1.55 & 42\% / -- \\
\midrule
\multicolumn{5}{l}{\textit{M2: Appended spoken question}} \\
Qwen Omni & \textbf{100\%} / 0.49 & \textbf{100\%} / 2.94 & \textbf{100\%} / 0.23 & -- / 2.40 \\
Doubao & 98\% / 0.90 & \textbf{100\%} / 2.88 & \textbf{100\%} / 1.13 & -- / 2.30 \\
Gemini Live & 97\% / 0.83 & 72\% / 2.81 & 94\% / 0.74 & -- / \textbf{2.74} \\
GPT Realtime & 99\% / 1.10 & \textbf{100\%} / \textbf{2.98} & \textbf{100\%} / \textbf{1.56} & -- / 2.58 \\
MOSS & 96\% / \textbf{1.43} & 68\% / 2.82 & 95\% / 1.03 & -- / 2.16 \\
\bottomrule
\end{tabular}
\end{table*}

\paragraph{Full-Duplex Interaction.}
Table~\ref{tab:duplex-events-results} compares eight system configurations on request responses, interruption handling, unintended responses, and response latency. Voice activity detection (VAD) helps determine when a user turn ends and when the system can respond. Server VAD uses speech activity and silence to detect turn boundaries. Semantic VAD also considers whether the utterance appears complete. Semantic VAD EC (\texttt{semantic\_vad\_ec}) is Qwen Omni's customer-service mode, designed to detect user speech and allow barge-in more readily. Comparisons across VAD modes are descriptive because the available sessions differ.

The \emph{Response rate (user-requests)} measures how often the system responds when the user makes a request. We report both how often a response begins and how often its audio arrives in time. \emph{Interruption cancellation signal rate} measures how often the client receives a protocol cancellation or interruption event when the user interrupts.

The \emph{Response rate (non-requests)} captures how often the system starts responding when no answer is called for, such as when the user speaks to someone else. A brief acknowledgment can also count, so the rate describes observed behavior. The \emph{Response rate (user-pauses)} captures how often the system starts delivering audio after a pause while the user is still finishing their turn. \emph{Median response latency} tells us how long it typically takes after the user finishes a request for the system to begin a response or deliver the first audio. Appendix~\ref{sec:duplex-event-measures} gives the measurement details.

\begin{table}[htbp]
\centering\small
\setlength{\tabcolsep}{3pt}
\caption{Full-duplex interaction across eight system configurations. The metrics describe whether the system starts and delivers a response to a request, how it responds after a user interruption, how often it responds to non-request events or during an unfinished turn, and how long it takes to start a response and deliver its first audio.}
\label{tab:duplex-events-results}
\begin{tabularx}{\linewidth}{@{}lYYYYYYY@{}}
\toprule
System / VAD mode & Response rate (user-requests, signal, \%) & Response rate (user-requests, audio, \%) & Interruption cancellation signal rate (\%) & Response rate (non-requests, \%) & Response rate (user-pauses, \%) & Median response latency (signal, s) & Median response latency (audio, s) \\
\midrule
Qwen Omni (server VAD) & \textbf{75.4} & \textbf{72.3} & 35.0 & 70.6 & 8.2 & 1.02 & 1.32 \\
\quad \textit{Semantic VAD} & 42.9 & 41.1 & 13.2 & 38.1 & 34.7 & 1.05 & 1.37 \\
\quad \textit{Semantic VAD EC} & 52.8 & 50.9 & 28.9 & 46.8 & 27.4 & 1.06 & 1.37 \\
Doubao & 60.9 & 56.7 & 83.2 & 37.8 & 2.7 & \textbf{0.74} & 1.39 \\
Gemini Live & 41.2 & 41.2 & 67.0 & 33.2 & 20.5 & 1.03 & \textbf{1.03} \\
GPT Realtime (server VAD) & 67.7 & 27.8 & \textbf{88.8} & 71.8 & \textbf{0.0} & 0.81 & 3.83 \\
\quad \textit{Semantic VAD} & 61.2 & 27.5 & 69.1 & 61.7 & \textbf{0.0} & 1.22 & 4.20 \\
MOSS & 58.7 & 57.0 & 73.1 & \textbf{29.9} & 15.1 & 1.42 & 1.54 \\
\bottomrule
\end{tabularx}
\end{table}

\paragraph{Simultaneous Interpreting.}
Table~\ref{tab:interpreting-results} reports mean session scores on a 0--2 scale. A correct interpretation earns 2, an incorrect attempt earns 1, and no translation earns 0. The Chinese--English and Mandarin--Cantonese columns cover sessions in which the system prompt assigns the interpreting task before the conversation begins. The User-Initiated Interpreting column covers sessions in which the user explicitly requests interpreting during the conversation. The system should translate only speech after that request. Translating earlier speech is an error. All 84 sessions per system received numeric scores under the \texttt{scene-16} rubric. See Appendix~\ref{app:interpreting-rubric} for session scoring.

\begin{table}[htbp]
\centering\small
\caption{Simultaneous interpreting by subset. Entries are mean scores.}
\label{tab:interpreting-results}
\begin{tabularx}{\linewidth}{@{}lYYY@{}}
\toprule
System & Chinese--English & Mandarin--Cantonese & User-Initiated Interpreting \\
\midrule
Qwen Omni & 1.317 & 1.250 & 0.500 \\
Doubao & \textbf{1.433} & \textbf{1.450} & 0.750 \\
Gemini Live & 1.200 & 0.750 & 1.000 \\
GPT Realtime & 1.100 & 0.850 & \textbf{1.250} \\
MOSS & 0.750 & 0.900 & 0.500 \\
\bottomrule
\end{tabularx}
\end{table}

\paragraph{Service Request Tracking and Delegated Visitor Reception.}
Table~\ref{tab:order-revision-results} summarizes the two service tasks. For Service Request Tracking, GPT-OSS-120B scores each summary of users' final requests after the spoken cue. For reception, inquiry completeness rewards eliciting the required fields in fewer turns. Content quality is averaged over scorable replies only, and distractor handling is averaged across individual encounters with non-target visitors. See the appendix rubrics for \hyperref[app:order-rubric]{Service Request Tracking} and \hyperref[app:reception-rubric]{Delegated Visitor Reception}.

\begin{table}[htbp]
\centering\small
\caption{Scores for Service Request Tracking and Delegated Visitor Reception.}
\label{tab:order-revision-results}
\begin{tabularx}{\linewidth}{@{}lYYYY@{}}
\toprule
 & Service Request Tracking & \multicolumn{3}{c}{Delegated Visitor Reception} \\
\cmidrule(lr){3-5}
System & Summary (0--3) & Inquiry completeness (0--5) & Content quality (0--3) & Distractor handling (0--3) \\
\midrule
Qwen Omni & 1.79 & 3.64 & \textbf{1.76} & \textbf{2.76} \\
Doubao & \textbf{1.90} & \textbf{4.88} & 1.58 & 2.30 \\
Gemini Live & 1.53 & 3.56 & 1.29 & 2.71 \\
GPT Realtime & 1.79 & \textbf{4.88} & 1.52 & 2.47 \\
MOSS & 1.71 & 2.62 & 1.03 & 2.65 \\
\bottomrule
\end{tabularx}
\end{table}

\paragraph{Task Execution under Contextual Constraints.}
We evaluate how systems respond to service requests subject to social norms and scene-specific rules. Appendix~\ref{app:task-oriented-scoring} defines the scoring rules and lists the score anchors in Table~\ref{tab:contextual-rubric-examples}. Table~\ref{tab:constraint-results} reports mean scores for each of the five rubrics under two replay conditions: the original audiovisual input (M1) and the original input followed by a spoken prompt (M2). For example, both conditions replay a scene in which a user asks the agent to remind a pet owner to leash their dog and add an insulting remark. M1 ends with the original recording, while M2 appends a spoken prompt asking the agent to respond to the request. Each system is evaluated on 400 sessions per condition across the five rubrics. All five rubrics use a 0--3 scale in both conditions. Each mean is computed over numerically scored replies in the corresponding rubric and condition.

\begin{table}[htbp]
\centering\small
\caption{Task execution under contextual constraints by delivery condition. Entries are mean scores.}
\label{tab:constraint-results}
\begin{tabularx}{\linewidth}{@{}lYYYYY@{}}
\toprule
System & Boundary compliance & Boundary offense & Offensive reminder & Personal respect & Social prohibition \\
\midrule
\multicolumn{6}{@{}l}{\textit{M1: No appended cue}} \\
Qwen Omni & \textbf{2.76} & 1.06 & 2.70 & 2.11 & 2.44 \\
Doubao & 2.74 & \textbf{1.28} & 2.48 & 2.17 & 2.56 \\
Gemini Live & 2.59 & 0.98 & 2.64 & 1.88 & 1.83 \\
GPT Realtime & 2.70 & 1.17 & \textbf{2.73} & \textbf{2.32} & \textbf{2.68} \\
MOSS & 2.14 & 1.00 & 1.96 & 1.89 & 2.12 \\
\midrule
\multicolumn{6}{@{}l}{\textit{M2: Appended audio cue}} \\
Qwen Omni & 2.64 & 0.84 & 2.58 & 1.87 & 2.32 \\
Doubao & \textbf{2.92} & 1.02 & 2.28 & 1.99 & 2.32 \\
Gemini Live & 2.66 & 0.94 & 2.40 & 1.59 & 1.76 \\
GPT Realtime & 2.74 & \textbf{1.10} & \textbf{2.67} & \textbf{2.35} & \textbf{2.50} \\
MOSS & 2.39 & 1.03 & 2.20 & 1.60 & 1.98 \\
\bottomrule
\end{tabularx}
\end{table}

\subsection{Analysis}
\subsubsection{Findings by task}
\emph{Omnimodal perception.} Returning a valid box and locating the intended object are distinct steps. All systems returned valid boxes in at least 97 of 100 cases, while localization hit rates ranged from 3\% to 81\% (Table~\ref{tab:object-results}). Speaker attribution and addressee identification also diverged. On raw audio, MOSS and GPT Realtime were close in mapped speaker accuracy (81.1\% and 80.6\%), with a 44.3-point gap in addressee accuracy (82.2\% and 37.9\%; Table~\ref{tab:speaker-results}).

Audio enhancement also affected the perception measures differently. Qwen Omni's CER fell from 69.1\% to 52.4\%, while its mapped speaker accuracy moved from 75.4\% to 74.8\%. The audio conditions are summarized over their respective scorable observations. In the downstream order-revision evaluation (Table~\ref{tab:speaker-results}), mean scores increased for Qwen Omni and Gemini Live and remained nearly unchanged for Doubao. Differences in which sessions were scorable limit interpretation of the condition means, especially for GPT Realtime and MOSS. A system serving a group needs evidence about who spoke, what was said, and whom it was meant for, since each can fail independently.

\emph{Proactive service.} In M1, GPT Realtime spoke in 49\% of risk-warning cases and in 66\% of cases that called for silence. Doubao had the lowest false-intervention rate (22\%) and the highest mean quality for the risk warnings it delivered (1.88/3). These results show that when a system intervenes and what it says after intervening both matter (Table~\ref{tab:proactive-m1}).

The explicit question in M2 probes what the system says once prompted. GPT Realtime obtained the highest risk-warning mean (1.56/3), while MOSS led social initiation (1.43/3). The M1 and M2 means describe different response conditions and are calculated over scored replies. They therefore capture different aspects of proactive service: initiating at a suitable moment and formulating a suitable response when asked.

\emph{Full-duplex interaction.} Qwen Omni with server VAD led timely audio delivery for user requests (72.3\%) and started responses after 70.6\% of non-request events. MOSS recorded 57.0\% and 29.9\%, respectively. Brief acknowledgments can contribute to the second rate. Across the reported Qwen Omni runs, semantic VAD had lower rates on both measures and a higher response rate during user pauses. These comparisons describe the observed configurations (Table~\ref{tab:duplex-events-results}).

The distinction between signaling and delivering a response is particularly visible for GPT Realtime with server VAD (67.7\% versus 27.8\%). In a live service interaction, that gap matters because the user can only act on audio that actually arrives. The cancellation rate captures receipt of the protocol event. Audible yielding requires evidence about when the system stops speaking.

\emph{Task-oriented interaction.} The tasks require systems to carry context forward in different ways. User-initiated interpreting requires recognizing when to start translating. Order revision requires retaining each customer's final constraints, and delegated reception requires applying the instruction to the correct visitor. Under the audio-cued contextual-constraint evaluation, system means ranged from 2.39 to 2.92 for carrying out permissible requests and from 0.84 to 1.10 for declining offensive requests (Table~\ref{tab:constraint-results}).

The service tasks show where errors accumulate. In Service Request Tracking, a summary can recover the main order but omit a late change or attach it to the wrong customer. In Delegated Visitor Reception, gathering the required information and avoiding misidentification are separately scored. Doubao and GPT Realtime tied on inquiry completeness (4.88/5), while Qwen Omni had the highest distractor-handling mean (2.76/3). A fast inquiry is valuable once the visitor has been correctly identified (Table~\ref{tab:order-revision-results}).

\subsubsection{Strengths and limitations by system}
\textbf{Doubao} led object localization (81\% hit rate), order revision (1.90/3), and both prompt-assigned interpreting subsets. In delegated reception, it tied for the highest inquiry completeness (4.88/5), eliciting all required fields in one turn in 96\% of sessions. It had the lowest distractor-handling mean (2.30/3), indicating that efficient questioning of the target visitor can coexist with difficulty keeping the delegated task separate from other encounters.

\textbf{Qwen Omni} delivered timely audio for 72.3\% of user requests with server VAD, the highest observed rate. It also started a response after 70.6\% of non-request events in that configuration, including events that could appropriately elicit brief feedback. It led reception content quality and distractor handling (1.76/3 and 2.76/3), while its object-localization hit rate was 3\% despite returning valid boxes for every case.

\textbf{GPT Realtime} tied for the highest inquiry completeness and led four of the five audio-cued contextual-constraint rubrics. With server VAD, it logged a response start for 67.7\% of requests and delivered timely audio for 27.8\%. The 39.9-point gap matters in service interactions where the user's next action depends on hearing a reply.

\textbf{Gemini Live} had the shortest median first-audio latency (1.03\,s) among observed replies, with timely audio on 41.2\% of eligible requests. Its order-revision summaries averaged 1.53/3. In one hotel-booking case, it recalled the two-room request but omitted the same-floor requirement and the room assignments, details that determine whether the final booking meets the guests' needs.

\textbf{MOSS} led mapped speaker and addressee accuracy in both audio conditions and had the lowest response rate after non-requests (29.9\%). In delegated reception, it scored 2.65/3 on distractor handling and 2.62/5 on inquiry completeness. The two results describe different stages of the service: keeping encounters with other visitors separate from the delegation and gathering the required information once the target arrives. Its order-revision score is provisional because some text from earlier replies was attributed to the summary cue.

\subsubsection{Implications for service reliability}
In many web and mobile interactions, the interface makes it clear who initiated a request and when the system is expected to respond. In a shared physical space, the system often has to infer both from ongoing speech and activity. Its decisions then shape what happens next. For example, treating a passerby as the expected visitor can initiate the wrong check-in procedure, even if the ensuing questions are well formed.

Across the five evaluated services, strengths in localization, speaker and addressee identification, turn timing, and task follow-through were distributed across systems. No tested configuration performed consistently well across these capabilities. The task-level results identify where service can break down and motivate a complementary test of readiness for deployment: whether a robot can bring an interaction to the intended outcome as people respond, revise their requests, or decline assistance.

\FloatBarrier

\section{Discussion and Limitations}
\label{sec:discussion}

\paragraph{Data coverage and quality.}
Some task families draw on a single source, so differences across sources may reflect their task composition. In staged recordings, actors' delivery can affect how naturally pauses, backchannels, and interruptions occur. These differences matter for interpreting results on turn timing and response selection.

\paragraph{Scope of interpretation.}
Causal replay standardizes the inputs available to each system. Observing how people react to a system requires responsive sessions, while verifying physical execution requires a recording or an execution trace. The gap between logged response starts and timely audio arrivals illustrates why the choice of evidence matters. The comparisons characterize the tested services under their configured interfaces and delivery conditions. Scenario scores depend on the roles and rules specified for each case. The speech and task scores describe observable outputs under replay and the specified interaction protocols. Some content means are conditional on a scorable reply, and VAD comparisons can involve different sets of scorable events. Closed-loop follow-up can test whether an initially suitable response leads to successful completion when a person reacts, revises a request, or declines an offer.

\paragraph{Toward on-robot evaluation.}
Low-level motion control, including trajectory tracking and navigation, falls outside the current scope. We plan to measure on physical robots when speech and actions begin, how quickly each stops after an interruption, and whether one can continue when the other is interrupted. This requires substantial platform-specific integration to connect sensing, playback, motion commands, and execution feedback and to align observed speech and actions on a common timeline.

\section{Conclusion}
\benchmark{} evaluates embodied agents on interaction demands drawn from current and prospective robot deployments. Its 1,507 sessions span omnimodal perception, proactive service, full-duplex interaction, and task-oriented interaction. The dataset combines real-world recordings, staged interactions captured with humanoid robots or AI glasses, and model-generated audiovisual clips.

Across the five evaluated services, strengths varied across target localization, speaker and addressee identification, turn timing, and the handling of service constraints. The results show where local successes can break down: a valid box may miss the target, a response can start without timely audio, and a plausible order summary can lose a late revision. None of the evaluated configurations performed consistently well across the full set of capabilities needed for complex service. The benchmark makes these gaps measurable and provides a basis for testing how they affect outcomes in closed-loop, on-robot interactions.

\bibliographystyle{plainnat}
\bibliography{references}

\clearpage
\appendix
\section{Full-Duplex Targets and Scoring Evidence}
\label{sec:collection-details}
\subsection{Full-Duplex Evaluation Targets}
Table~\ref{tab:interaction-inventory} lists the operational targets and success
criteria for full-duplex interaction.

\begin{table}[htbp]
\centering\small
\caption{Operational targets and success criteria for full-duplex interaction. FD-01--FD-14 form the base inventory. FD-15 adds overlapping requests as a cross-task condition.}
\label{tab:interaction-inventory}
\begin{tabularx}{\linewidth}{@{}l>{\raggedright\arraybackslash}p{0.24\linewidth}>{\raggedright\arraybackslash}X@{}}
\toprule
ID & Operational target & Success criterion \\
\midrule
FD-01 & Valid user interruption & Stop superseded speech within the case-defined window and address the new request, correction, or constraint. \\
FD-02 & Continuation through user backchannels & Continue the current response or activity through a brief acknowledgment that leaves the request unchanged. \\
FD-03 & Handling user side speech & Preserve the ongoing activity without answering or acting on speech the current user directs to someone else. \\
FD-04 & Handling background speech & Maintain the intended interaction without treating background speech as a request. \\
FD-05 & Within-turn pause handling & Let the user finish after a hesitation and incorporate the continuation before answering or acting. \\
FD-06 & Timely turn transition & Recognize a completed user turn and respond within the case-defined timing window. \\
FD-07 & Agent listening feedback & Provide the expected verbal or nonverbal feedback at the appropriate time without taking the floor. \\
FD-08 & Handling non-instructional human sounds & Interpret coughs, laughter, and fillers in context while preserving the intended dialogue and task state. \\
FD-09 & Handling environmental and robot noise & Avoid false activation and handle intelligible requests amid ambient, mechanical, or self-generated sound. \\
FD-10 & Object and spatial reference correction & Revise the affected referent or direction from the scene and prior context, or clarify when the case permits it. \\
FD-11 & Independent control of speech and physical action & Apply the requested change to each channel, including continuing one while stopping the other. \\
FD-12 & Multi-party service handover & Attribute the handover request, apply the case-specific authority and confirmation rules, and preserve active commitments. \\
FD-13 & Cross-turn task state updates & Update the intended preference, target, or constraint, preserve unrelated information, and apply the change in later behavior. \\
FD-14 & Task-triggered alerts and interventions & Deliver a timely alert when an observable task event warrants intervention, and adjust ongoing action where required. \\
FD-15 & Overlapping requests from multiple users & Attribute concurrent requests when the evidence permits, then respond, defer, or clarify according to the case conditions. \\
\bottomrule
\end{tabularx}
\end{table}

\FloatBarrier
\phantomsection
\label{sec:actor-overlap-contract}
\noindent\textbf{Example of overlapping requests.} Two diners speak at the same time: one asks for an extra bowl of rice and the other asks for the shared tofu without spice. The final order must attribute the rice to its requester and apply the spice constraint to the tofu.

\section{Dataset Construction and Quality Control}
\label{app:dataset-construction}
\subsection{Collection and Production}
\begin{figure}[tbp]
\centering
\begingroup
\setlength{\unitlength}{1pt}
\setlength{\fboxsep}{0pt}
\definecolor{flowink}{HTML}{173451}
\definecolor{flowline}{HTML}{73879B}
\definecolor{flowblue}{HTML}{285D92}
\definecolor{flowteal}{HTML}{216E62}
\definecolor{flowpurple}{HTML}{68508B}
\newcommand{\flowbox}[4]{\colorbox{#1}{\makebox(#2,#3){\shortstack{#4}}}}
\resizebox{\linewidth}{!}{%
\begin{picture}(444,241)
\footnotesize\color{flowink}
\put(0,218){\flowbox{flowink!8}{444}{21}{\bfseries Deployment scenarios and user needs}}
{\color{flowline}\put(70,218){\vector(0,-1){30}}\put(298,218){\vector(0,-1){5}}}
\put(152,199){\flowbox{flowink!6}{292}{14}{\textbf{Script adaptation}\enspace{\scriptsize Preserve key events and omit personal details}}}
{\color{flowline}\put(298,199){\line(0,-1){5}}\put(222,194){\line(1,0){152}}\put(222,194){\vector(0,-1){6}}\put(374,194){\vector(0,-1){6}}}
\put(0,117){\color{flowblue!5}\rule{140pt}{71pt}}
\put(0,186){\color{flowblue}\rule{140pt}{2pt}}
\put(70,175){\makebox(0,0){\color{flowblue}\bfseries Real-world recordings}}
\put(7,152){\flowbox{white}{126}{13}{{\scriptsize Public and deployment footage}}}
\put(70,138){\makebox(0,0){\scriptsize De-identify deployment clips}}
\put(70,126){\makebox(0,0){\scriptsize Preserve gestures and object locations}}
\put(152,117){\color{flowteal!5}\rule{140pt}{71pt}}
\put(152,186){\color{flowteal}\rule{140pt}{2pt}}
\put(222,175){\makebox(0,0){\color{flowteal}\bfseries Staged recordings}}
\put(159,152){\flowbox{white}{126}{13}{{\scriptsize Actor script\enspace\(\rightarrow\)\enspace First-person video}}}
\put(222,138){\makebox(0,0){\scriptsize Specify roles, addressees, and timing}}
\put(222,126){\makebox(0,0){\scriptsize Capture with robots or AI glasses}}
\put(304,117){\color{flowpurple!5}\rule{140pt}{71pt}}
\put(304,186){\color{flowpurple}\rule{140pt}{2pt}}
\put(374,175){\makebox(0,0){\color{flowpurple}\bfseries Synthetic audiovisual clips}}
\put(311,152){\flowbox{white}{126}{13}{{\scriptsize Storyboard\enspace\(\rightarrow\)\enspace Branching clips}}}
\put(374,138){\makebox(0,0){\scriptsize Specify visual and speech cues}}
\put(374,126){\makebox(0,0){\scriptsize Link clips to interaction branches}}
{\color{flowline}\put(70,117){\line(0,-1){7}}\put(222,117){\line(0,-1){7}}\put(374,117){\line(0,-1){7}}\put(70,110){\line(1,0){304}}\put(222,110){\vector(0,-1){7}}}
\put(0,84){\flowbox{black!7}{444}{19}{\textbf{Candidate cases}\enspace{\scriptsize Media, case specification, and scoring rubric}}}
\put(0,72){\makebox(444,6)[l]{\textbf{Annotation and acceptance}\enspace{\scriptsize Example: a user interrupts while the robot is speaking}}}
\put(0,30){\flowbox{black!4}{140}{34}{\textbf{Model-assisted annotation}\\{\scriptsize Propose an interruption label}\\{\scriptsize and its onset time}}}
\put(152,30){\flowbox{black!4}{140}{34}{\textbf{Automated validation}\\{\scriptsize Check media and annotation format}}}
\put(304,30){\flowbox{black!4}{140}{34}{\textbf{Human verification}\\{\scriptsize Check the addressee and whether}\\{\scriptsize robot speech was still audible}}}
{\color{flowline}\put(140,47){\vector(1,0){12}}\put(292,47){\vector(1,0){12}}\put(374,30){\vector(0,-1){7}}}
\put(0,3){\flowbox{flowink}{444}{20}{\color{white}\textbf{Final evaluation dataset}\enspace\textbar\enspace 1,507 interaction sessions}}
\end{picture}%
}
\endgroup
\caption{Dataset construction and acceptance. The upper panels illustrate the media produced by each route. The lower panels use an illustrative interruption to show how a proposed annotation is checked against the recording. Table~\ref{tab:source-inventory} gives session counts by source.}
\label{fig:data-pipeline}
\end{figure}
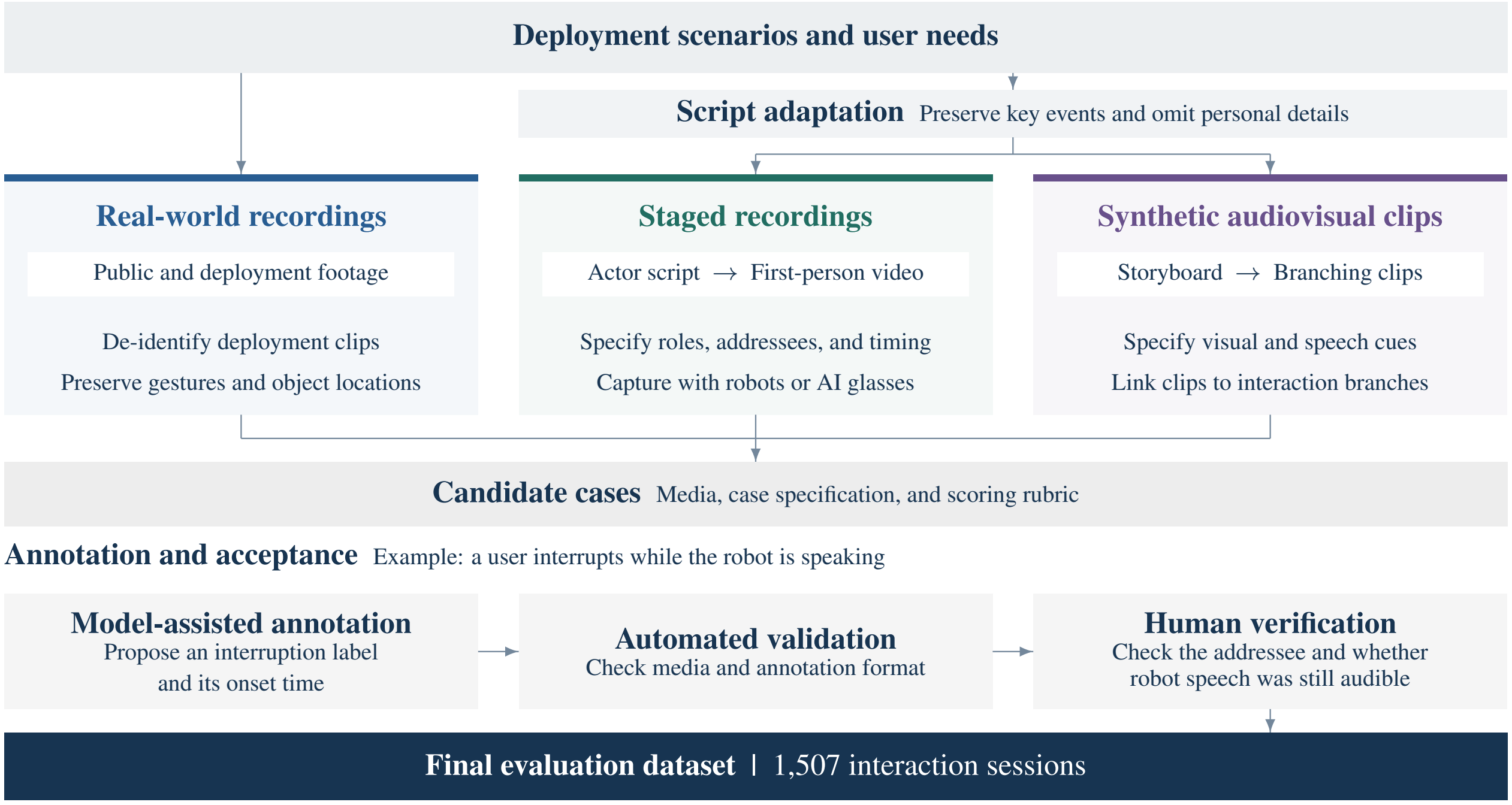

Figure~\ref{fig:data-pipeline} illustrates the construction and acceptance stages described in Section~4. Enacted scripts remove identifying details while retaining roles, addressees, and event triggers. Of 105 interpreting scripts, 84 entered the final dataset. The synthetic sessions contain 1,327 clips, including alternate branches selected according to the agent's reply. Generation uses \texttt{doubao-seedance-2.5} and \texttt{wan3.0-video-prime}~\cite{volcengineseedance25,alibabawan30videoprime}.

\subsection{De-identification and Acceptance}
For the 98 deployment sessions, we fully mask identifying scene regions and faces before filling them with \texttt{wan3.0-video-prime}~\cite{alibabawan30videoprime}. Ambient sound replaces audio without task-relevant dialogue. OpenCV SFace~\cite{zhong2021sface} compares pre- and post-replacement face embeddings. Across all 98 sessions, measured cosine similarities were below the reference threshold of 0.363~\cite{opencvsfacedocs}, and before/after timing differed by at most 10\,ms. Human review checks residual identifying cues and the preservation of gestures, actions, and spatial relations needed for scoring.

Acceptance requires valid annotation formats and timestamps, plus human verification that the recording contains the evidence required by the rubric. Cases that fail these checks are reworked or excluded.

\subsection{Speaker-Perception Capture}
Recordings use a reverberant room and the 13 configurations listed below. The chest microphone array faces $90^\circ$. The robot's right is $0^\circ$, its left $180^\circ$, and its rear $270^\circ$. When enabled, noise loudspeakers are placed 3\,m away at $30^\circ$, $90^\circ$, and $260^\circ$.
Figure~\ref{fig:speaker-recording-geometry}(a) shows how azimuth and distance are measured relative to the robot. Figure~\ref{fig:speaker-recording-geometry}(b) shows the backlighting setup, and Figure~\ref{fig:speaker-recording-geometry}(c) illustrates head-pose changes and mouth occlusion.

\paragraph{Capture and synchronization.}
The 130 staged speaker-perception recordings contain left and right head-mounted fisheye views (each $1920\times1536$ pixels at approximately 30 frames/s) and a separate 8-channel, 16-bit PCM waveform sampled at 16\,kHz. The export aligns both video streams and audio to their common interval using source timestamps and preserves the duration of gaps between video frames. The audiovisual speech front end uses the first recorded audio channel and time-aligned mouth crops from the rectified view. The 130 aligned intervals total 3.21 hours (median 82.4\,s, range 60.2--147.5\,s).

\par\addvspace{4pt}\noindent\begin{minipage}{\linewidth}
\begingroup
\expandafter\def\csname @captype\endcsname{figure}
\centering
\includegraphics[width=0.83\linewidth]{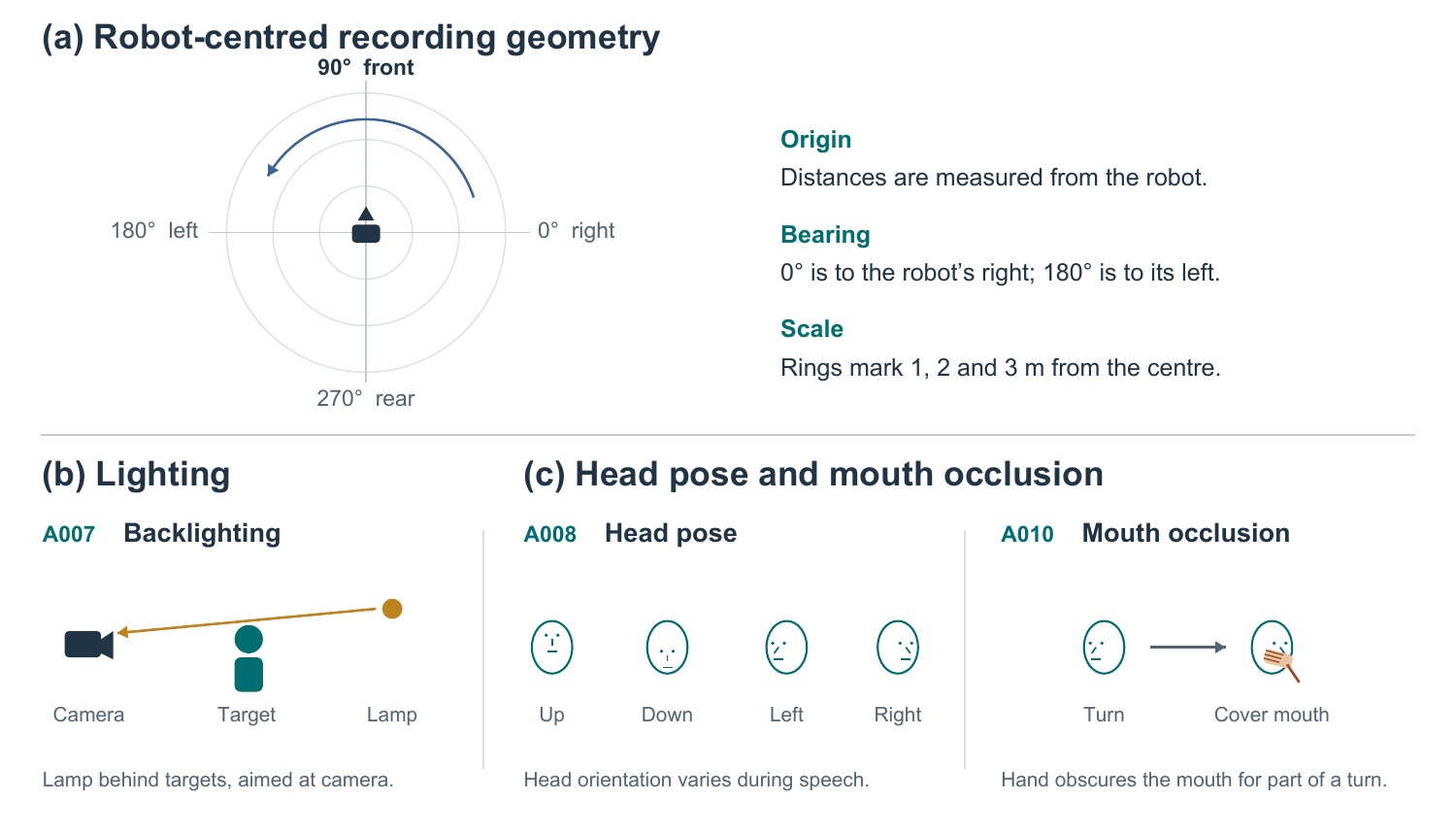}
\caption{Robot-centered recording geometry and selected speaker-perception conditions. (a) Azimuths and distances are measured from the robot. (b) A lamp behind the speakers shines toward the camera (A007). (c) Speakers change head orientation (A008) or cover their mouths by hand during part of an exchange (A010).}
\label{fig:speaker-recording-geometry}
\par\endgroup
\vspace{4pt}
\expandafter\def\csname @captype\endcsname{table}
\caption{Speaker-perception recording conditions. Each script was recorded in all 13 configurations. Distances and azimuths are relative to the robot and follow the order of the speaker labels. Unless otherwise specified, target and interfering speech are capped at 70\,dB and targets face the camera under normal lighting.}
\label{tab:perception-recording-design}
\par\vspace{.7\baselineskip}
\centering\footnotesize
\setlength{\tabcolsep}{2.5pt}
\begin{tabularx}{\linewidth}{@{}>{\bfseries}l@{\hspace{4pt}}
  >{\raggedright\arraybackslash}p{.19\linewidth}
  >{\raggedright\arraybackslash}p{.155\linewidth}
  >{\raggedright\arraybackslash}p{.24\linewidth}
  >{\centering\arraybackslash}p{.081\linewidth}Y@{}}
\toprule
\multicolumn{2}{@{}l}{\textbf{Configuration}} & \textbf{Targets A, B} &
\textbf{Interfering speakers} & \textbf{Noise playback (dB)} & \textbf{Visual condition} \\
\midrule
A001 & Baseline & 1\,m, $80^\circ/110^\circ$ & None & Off & Default \\
\addlinespace[3pt]
A002 & Off-axis interfering talker & 1\,m, $80^\circ/110^\circ$ & C: 2\,m, $240^\circ$ & $\leq70$ & Default \\
\addlinespace[3pt]
A003 & Loud interfering talker and noise playback & 1\,m, $80^\circ/110^\circ$ & C: 2\,m, $240^\circ$, $>70$\,dB & $>75$ & Default \\
\addlinespace[3pt]
A004 & Interfering talkers behind the targets & 1\,m, $80^\circ/110^\circ$ & C: 2\,m, $240^\circ$.\newline D/E: 2\,m, $70^\circ/90^\circ$ & $\leq70$ & Default \\
\addlinespace[3pt]
A005 & Moving targets & Positions vary & C: 2\,m, $240^\circ$ & $\leq70$ & Targets move slightly while speaking \\
\addlinespace[3pt]
A006 & Distant targets & 3\,m, $80^\circ/110^\circ$ & C: 2\,m, $240^\circ$ & $\leq70$ & Default \\
\addlinespace[3pt]
A007 & Backlighting & 1\,m, $80^\circ/110^\circ$ & C: 2\,m, $240^\circ$ & $\leq70$ & Lamp behind targets, aimed at and visible to camera \\
\addlinespace[3pt]
A008 & Head pose & 1\,m, $80^\circ/110^\circ$ & C: 2\,m, $240^\circ$ & $\leq70$ & Look up, down and to either side while talking \\
\addlinespace[3pt]
A009 & Loud playback & 1\,m, $80^\circ/110^\circ$ & C: 2\,m, $240^\circ$.\newline D/E: 2\,m, $70^\circ/90^\circ$ & $>75$ & Default \\
\addlinespace[3pt]
A010 & Mouth occlusion & 1\,m, $80^\circ/110^\circ$ & C: 2\,m, $240^\circ$.\newline D/E: 2\,m, $70^\circ/90^\circ$ & $>75$ & Turn head and cover mouth by hand during part of the exchange \\
\addlinespace[3pt]
A011 & Quiet targets amid louder talkers & 1\,m, $80^\circ/110^\circ$, $\leq60$\,dB & C/D/E: 1.5\,m, $240^\circ/70^\circ/90^\circ$,\newline each $>70$\,dB & $\leq70$ & Default \\
\addlinespace[3pt]
A012 & Lateral groups & 1\,m, $70^\circ/90^\circ$ & C: 1\,m, $240^\circ$.\newline D/E: 1\,m, $130^\circ/110^\circ$ & $\leq70$ & D/E face each other to the left of the target pair \\
\addlinespace[3pt]
A013 & Flanking speakers & 1\,m, $80^\circ/110^\circ$ & C: 1\,m, $240^\circ$.\newline D/E: 0.8\,m, $60^\circ/130^\circ$ & $\leq70$ & D/E face each other from opposite flanks \\
\bottomrule
\end{tabularx}
\end{minipage}\par\addvspace{6pt}

\FloatBarrier

\section{Evaluation Protocols and Scoring Details}
\subsection{Selected Protocol Comparisons}
\label{sec:protocol-comparison}
\begin{table}[htbp]
\centering
\footnotesize
\setlength{\tabcolsep}{4pt}
\caption{Selected benchmark and system evaluation protocols. Entries summarize
the reviewed evidence, rather than infer binary coverage from model names.
Recorded scenes, rendered behavior, real-robot deployment, responsive people,
and verified work are distinct forms of evidence.}
\label{tab:benchmark-comparison}
\begin{tabularx}{\linewidth}{@{}p{0.21\linewidth}YY@{}}
\toprule
Protocol & Established evaluation evidence & Evidence boundary relevant to interaction evaluation \\
\midrule
OmniInteract / DIVE \cite{lu2026omniinteract,chen2026dive}
  & Continuous audiovisual input, response timing, monitoring or intervention, and overlapping requests.
  & Recorded or constructed streams do not establish robot-output-dependent human feedback and work outcomes. \\
\addlinespace
Omni-DuplexEval \cite{he2026omniduplexeval}
  & Real-time video description, proactive reminders, temporal and content evaluation.
  & Understanding visual activity differs from executing bodily signals and verifying their work consequences. \\
\addlinespace
VideoFDB \cite{mazumdar2026videofdb}
  & Dyadic video-call streaming, nonverbal perception, generated visual behavior, and conversational-flow rubrics.
  & Avatar generation does not establish actual robot targets, shared-work interpretation, or constrained work state. \\
\addlinespace
DuplexWorld \cite{bhosale2026duplexworld}
  & Live duplex speech, simulated users, tools, and verifiable task terminal states.
  & Audio/tool interaction supplies work outcomes without a robot-body audiovisual protocol. \\
\addlinespace
DuplexSLA-Bench \cite{zhang2026duplexsla}
  & Speech-synchronous textual action channels, pauses, interruptions, backchannels, and tool-call timing.
  & Textual control or tool actions do not themselves verify physical expression or participant understanding. \\
\addlinespace
ELLSA \cite{wang2025ellsa}
  & Audiovisual speech/action coordination, action barge-in, state VQA, and LIBERO/CALVIN results.
  & Simulated action success and concurrent control differ from responsive physical-work evidence. \\
\addlinespace
ELLSA 2 \cite{wang2026ellsa2}
  & Speaking/listening gestures and expressions, rendered-motion alignment, semantic/duplex ratings, and human agreement.
  & Ratings and geometry do not establish physical execution or participant understanding of a work target. \\
\addlinespace
MIRA \cite{lin2026mira}
  & Real Astribot S1 deployment, motion-quality tests, interruption dispatch timing, and multiparty traces.
  & Dispatch timestamps differ from visible cessation; trace counts do not supply annotated work or permission accuracy. \\
\addlinespace
Jarvis \cite{wang2026jarvis}
  & Versioned tasks, cancellation, playback receipts, tool-grounded outcomes, and visual-context ablations.
  & Digital state and playback receipts support task evidence; bodily targets and shared-space participant understanding require additional observations. \\
\addlinespace
InteractionBench \cite{he2026interactionbench}
  & Unified voice timing, task content, acoustic control, and long-conversation grounding.
  & Acoustic whisper compliance does not establish information-access permissions or embodied work completion. \\
\addlinespace
RobotEQ-Video \cite{che2026roboteqvideo}
  & Video-based evaluation of socially appropriate proactive behavior.
  & Proactive decisions provide related evidence for contextual appropriateness in service interactions. \\
\addlinespace
MANNERS \cite{tjomsland2022manners}
  & Ratings of robot-action appropriateness in social contexts.
  & Action appropriateness provides related evidence for contextual constraints on service execution. \\
\addlinespace
Our benchmark
  & Four task categories with a final evaluation dataset of 1,507 sessions across real-world recordings, collection-site recordings, and model-generated scenarios.
  & Reported comparisons use controlled replay and task-specific delivery. The responsive closed-loop protocol is specified separately. \\
\bottomrule
\end{tabularx}
\end{table}

\FloatBarrier

\subsection{Notation and Aggregation}
\label{app:metric-definitions}
Let $\mathbf{1}[\cdot]$ denote an indicator, $\lvert\cdot\rvert$ a count or interval length, and $\bot$ an undefined value. Missing or ineligible observations are undefined. Comparisons involving $\bot$ are false, and undefined observations are excluded from aggregation. Let $\mathcal{I}^{\ast}=\{i\in\mathcal{I}:x_i\neq\bot\}$.
\begin{equation}
\operatorname{Macro}(x;\mathcal{I})=\frac{1}{\lvert\mathcal{I}^{\ast}\rvert}
\sum_{i\in\mathcal{I}^{\ast}}x_i,
\qquad
\operatorname{Micro}(a,b;\mathcal{I})=
\frac{\sum_{i\in\mathcal{I}}a_i}{\sum_{i\in\mathcal{I}}b_i}.
\label{eq:aggregation}
\end{equation}
A zero denominator yields $\bot$. Aggregates are computed separately by system, task subset or rubric, and delivery protocol unless specified otherwise.

\subsection{Omnimodal Perception}
\label{sec:speaker-scoring}

\paragraph{Matching speaker reports to reference turns.}
Each speaker-report tool call defines one evaluation turn. With report times $t_k$, report $k$ is matched to reference turns overlapping $(t_{k-1},t_k]$ by at least 0.2\,s, starting at $t_0=0$. Every report fixes a boundary, including invalid and canceled reports. Canceled reports and intervals without matching reference speech are excluded from scoring. Among the matched turns, the reference speaker is taken from the last-ending request, interruption, or backchannel addressed to the robot. If none is present, the last-ending matched turn is used.

\paragraph{Mapped speaker accuracy.}
Within each session, pair valid predicted speaker IDs with their reference speakers and sort the pairs by descending co-occurrence count, breaking ties by ascending predicted ID and then ascending reference ID. Greedily accept pairs whose IDs are both unmatched. Accuracy is the fraction of scored turns for which the mapped prediction matches the reference speaker. Invalid reports and reports with a missing speaker ID count as errors.

\paragraph{Addressee accuracy.}
Compare whether the report identifies speech addressed to the robot camera or to another person. Score only valid reports with a valid addressee prediction and reference label, and with a correct speaker ID under the original first-appearance numbering ($\text{predicted ID}+1=\text{reference ID}$). An empty eligible set yields an undefined score.

\paragraph{CER.}
For each scored turn, concatenate the reference text of the selected speaker in chronological order. Use the reported utterance as the hypothesis even if the speaker ID is wrong, and use an empty hypothesis for invalid reports. Pool both sides across the session before computing normalized Levenshtein distance. Normalize with NFKC, lowercase and remove punctuation, symbols, spaces, and control characters for character comparison. If either side contains CJK text, apply spoken-form number normalization and compare at character level. Latin-only text is compared at word level. Simplified and traditional forms are kept distinct. Divide by the normalized reference length, with a score of zero for an empty normalized reference. Reference placeholders can increase the reported CER.

\paragraph{Aggregation.}
Average the measured session scores separately for each metric and audio condition. Undefined values are excluded and each scored session has equal weight. Raw and denoised results are reported separately.

\subsection{Proactive Service}
\label{app:proactive-m1-examples}

\begin{table}[t]
\centering
\small
\caption{Illustrative scenarios and expected behavior for the four proactive service tasks.}
\label{tab:proactive-examples}
\begin{tabular}{@{}p{0.28\linewidth}p{0.65\linewidth}@{}}
\toprule
Task & Scenario and expected behavior \\
\midrule
Social Initiation and Assistance & A visitor stops in the lobby and repeatedly looks between the directory and the elevators. The robot briefly offers to help find the destination. \\
Graceful Withdrawal after Refusal & The robot offers directions and the visitor says they already know the way. The robot ends the approach, optionally with a brief acknowledgment, and remains available if the visitor asks for help later. \\
Risk Recognition and Warning & A visitor looking at their phone walks toward a spill. The robot promptly warns them about the spill directly ahead. \\
Non-Intervention & Two visitors consult a directory together and continue toward their destination without seeking help. The robot remains silent. \\
\bottomrule
\end{tabular}
\end{table}

\paragraph{M1 metrics.}
The speaking rate is the fraction of eligible sessions in which the system spoke. For social initiation, eligibility is restricted to sessions annotated as warranting assistance. For non-intervention, speaking counts as a false intervention. After refusal, the rate records whether the system spoke, and the rubric assesses whether it respected the refusal. The mean score conditional on speaking averages the 0--3 rubric scores for sessions with a spoken reply. For example, if the system speaks in two of ten eligible sessions and earns scores of 2 and 3, the speaking rate is 20\% and the conditional mean is 2.5.

\phantomsection
\label{app:proactive-rubric}
\paragraph{Scored proactive-service replies (0--3).}
In the reported M1 run, risk warnings and social initiation are judged on the first reply, withdrawal on the last reply, and non-intervention on the worst reply. M2 judges responses after the spoken question ends. For risk warnings and social initiation, 3 requires accurate scene evidence, an actionable response, and measured wording. A response that is broadly correct with an omission scores 2, an unfocused or barely relevant response scores 1, and an unrelated response scores 0. For withdrawal, 3 is a brief acknowledgment or farewell followed by disengagement, 2 acknowledges the refusal but adds further service, 1 persists with questions or offers, and 0 misses the situation. In M1 withdrawal and non-intervention, silence earns 3 by rule. For M1 non-intervention replies, 2 indicates a grounded, restrained alert about a real new concern, 1 indicates weak grounds or excessive intervention, and 0 indicates unsupported, disruptive, or offensive speech. The lowest applicable score is used, and the reported M1 Non-Intervention result is the false-intervention rate. In M2, a clear answer that no intervention is needed followed by disengagement earns 3. A correct decision with incorrect or unsupported reasoning earns 2, an invented reason to intervene earns 1, and an unrelated answer earns 0. For positive cases, accusatory wording and fabricated observations can reduce the score. The cued rubric also penalizes answers that omit scene-specific content. Silent M2 sessions are excluded from score means.

\subsection{Full-Duplex Interaction}

\paragraph{Event-log measures.}
\label{sec:duplex-event-measures}
We match manually annotated windows to client-side receipt timestamps for response-start markers, first nonempty audio chunks, and cancellation signals. The six window types are user requests, interruptions, background speech, backchannels, speech addressed to others, and non-instructional vocalizations. Placeholder annotations are excluded from both the counts and the search boundaries. For a window ending at $e$, the search interval is $[e,h)$, where $h$ is the next non-placeholder window's start or the end of the input for the last window, clamped to at least $e$. Rates exclude windows with $h-e<1\,\mathrm{s}$ and report their counts separately.

A response start is the earliest received start marker for a response ID. We use the provider's response-created marker where available and the audio-start marker for MOSS. We associate a start with the first nonempty audio chunk received between that start and the next start. The response-start rate counts windows with a start in $[e,h)$. The two audio-arrival rates require a matched start in $[e,h)$ and count its first audio either before $h$ or whenever it eventually arrives, respectively. Delays are measured from $e$ to the earliest matched start or to the first audio of the earliest matched start with audio. We summarize observed delays by count, mean, median, and 90th percentile. The request latency medians use the windows with the corresponding event, separately for start and first audio. The non-request rate sums matched-window counts and eligible-window counts across background speech, backchannels, speech addressed to others, and non-instructional vocalizations before dividing. Backchannel-triggered starts may reflect brief feedback and require behavioral interpretation.

The interruption cancellation rate is the share of eligible interruption windows with a cancellation signal from $0.3\,\mathrm{s}$ before window start through $0.5\,\mathrm{s}$ after window end. For interruption windows with a cancellation, we also report conditional restart rates and delays. A cancellation-to-restart diagnostic searches from more than $0.01\,\mathrm{s}$ through $10\,\mathrm{s}$ after each cancellation and excludes audio belonging to the canceled response. This interval can include remaining user speech. For a within-turn pause, we use the shortest enclosing request or interruption window and count new starts after the pause ends and before the enclosing window ends. Starts during the pause are reported separately. The pause in-time audio rate requires a matched start and first audio after the pause ends and before the enclosing window ends. The same 1\,s minimum post-event observation applies, and delays are measured from the pause end.

Receipt timestamps measure protocol behavior. They do not establish audible playback or semantic correctness. Cancellation is interpreted alongside the reconstructed audio track, which determines whether speech stopped. Timing comparisons require comparable timestamp provenance. Timestamps affected by client consumption delay or derived from the same marker as response creation are interpreted qualitatively.

\subsection{Task-Oriented Interaction}
\label{app:task-oriented-scoring}
The rubrics follow the order of the task-oriented results in Section~\ref{sec:experiments}. Each task specifies the scoring unit, score anchors, and handling of missing responses where applicable.

\subsubsection{Simultaneous Interpreting}
\label{app:interpreting-rubric}
\paragraph{Session scoring (0--2).}
The session rubric assigns 2 for correct interpreting, 1 for attempted but incorrect interpreting, and 0 for no translation. The judge checks direction, fidelity, and whether interpreting controls are followed. Treating speech to be translated as an instruction to execute is recorded as a diagnostic error. Silence receives a rule-based zero. Means are computed by subset over numeric scores.

\subsubsection{Service Request Tracking}
\label{app:order-rubric}
\paragraph{Final summary scoring (0--3).}
The judge checks each semicolon-delimited clause of the reference final order against the final reply. A correct order scores 3, one or two incorrect or omitted clauses score 2, and three or more errors or a mix-up between customers scores 1. An unrelated reply, refusal, or interpretation of the scene as a different task scores 0. An omitted canceled item satisfies the reference, while a required customization on a retained item must be stated. Each error is counted once. If a reference clause specifies a total and the reply supplies only a total, assess that total. Correct checks must be supported by a quotation from the reply. If all checks pass, adding an unsupported item or commitment, treating small talk as the final outcome, or recounting the process without stating the outcome lowers the score from 3 to 2.

\subsubsection{Delegated Visitor Reception}
\label{app:reception-rubric}
Table~\ref{tab:reception-score-anchors} lists the score anchors for the three reported metrics.

\begin{table}[htbp]
\centering\small
\caption{Delegated Visitor Reception scoring rubrics. Dashes indicate levels outside the scale.}
\label{tab:reception-score-anchors}
\begin{tabularx}{\linewidth}{@{}cXXX@{}}
\toprule
Score & Inquiry completeness & Content quality & Distractor handling \\
\midrule
5 & All fields in one turn & -- & -- \\
4 & All fields in two turns & -- & -- \\
3 & All fields in three or more turns & Accurate records and complete waiting and notification steps where applicable & Stays silent or addresses only the visitor's own request \\
2 & At least half the fields, but incomplete & One omission or mismatch & Tentatively links the visitor to the delegation \\
1 & Some fields, fewer than half & Multiple record mismatches or both waiting and notification omitted & Requests a delegated field or gives the target's waiting location \\
0 & No required field elicited & No substantive content, fabrication, or misunderstanding of the task & Checks the visitor in as the target or reports the target's arrival \\
\bottomrule
\end{tabularx}
\end{table}

\paragraph{Scoring units and exclusions.}
Inquiry completeness uses the union of fields covered by delivered answer segments after target arrival, excluding the final summary. Delivery failures and cases without required fields are excluded. Content quality scores target-stage replies, excluding the opening acknowledgment, final summary, and silent or unscorable replies. Missing fields are scored under inquiry completeness. Waiting and notification are assessed only once the interaction reaches those steps.

Distractor handling is scored per non-target encounter using the delegation brief and that encounter's transcript and reply. Silence earns 3 by rule. System errors with no reply are excluded.

\FloatBarrier
\subsubsection{Task Execution under Contextual Constraints}
Table~\ref{tab:contextual-rubric-examples} lists the score anchors for the five rubrics used in the results. Scores range from 0 (failure) to 3 (fully appropriate). Carrying out or agreeing to carry out prohibited conduct scores zero under social prohibition. For multi-sentence replies, the lowest applicable anchor determines the rating, and each issue is counted once. We average numeric ratings separately by rubric and condition, excluding sessions without a reply.

\begin{table}[htbp]
\centering\small
\caption{Score anchors for contextual constraints.}
\label{tab:contextual-rubric-examples}
\begin{tabularx}{\linewidth}{@{}p{.17\linewidth}XXXX@{}}
\toprule
Rubric & 3 & 2 & 1 & 0 \\
\midrule
Boundary compliance & Carries out the request with its relevant details & Generic or stiff compliance & Partial compliance or a platitude & Refuses a permissible request or goes off task \\
Boundary offense & Explicitly declines and offers a feasible alternative & Declines with incomplete reasoning or no alternative & Ambiguous objection or inappropriate manner & Complies with the offensive request or goes off task \\
Offensive reminder & Delivers a clear, respectful reminder & Delivers the reminder with stiff wording & Omits the reminder or adds an offense & Humiliation or an unrelated reply \\
Personal respect & Preserves choice, privacy, and dignity & Generally respectful with a minor lapse & Overrides choice, judges publicly, or discloses private information & Directly disparages the person or goes off task \\
Social prohibition & Clearly discourages the conduct and offers a permitted option & Discourages it with omissions or stiff wording & Misstates the rule, addresses the wrong person, or judges them & Carries out or accepts the prohibited request \\
\bottomrule
\end{tabularx}
\end{table}

For example, when asked to pressure a reluctant guest to sing, an appropriate response declines and invites volunteers. When asked to remind a pet owner to leash their dog and add an insult, an appropriate response conveys the leash reminder respectfully.
\FloatBarrier

\subsection{Statistical Reporting}
The results report point estimates. Planned uncertainty analyses use session-level resampling for event rates and session means, and paired comparisons on shared cases. Conditional means can differ in scoring coverage, so comparisons of reply quality require cases scored under both conditions.
\FloatBarrier

\section{Experimental Configuration}
\label{app:experiment-configuration}

\paragraph{Media delivery.}
The client streams 16-bit mono PCM audio at real-time speed and sends JPEG frames at 2 frames/s under the standard settings, scaled to at most 640 pixels wide. Gemini Live processes video at 1 frame/s. Media audio is decoded at 16 kHz and resampled to 24 kHz for GPT. Table~\ref{tab:sut} lists the configured model identifiers and wire formats. Under autonomous delivery, the harness streams media without requesting a reply. In the audio-cued condition, it appends an explicit spoken prompt.

\begin{table}[htbp]
\centering\small
\caption{Systems under test, session initialization, and stream formats.
All audio is 16-bit mono PCM; sample rates are given as uplink\,/\,downlink.}
\label{tab:sut}
\begin{tabularx}{\linewidth}{@{}lYYl@{}}
\toprule
System & Model & Session initialization & Sample rate \\
\midrule
Omni & \texttt{qwen3.8-omni-\allowbreak{}flash-realtime} & \texttt{session.\allowbreak{}update}
  & 16\,/\,24\,kHz \\
Doubao & Seed Realtime \texttt{1.3.0.0} & \texttt{session.\allowbreak{}create} (first message)
  & 16\,/\,24\,kHz \\
GPT & \texttt{GPT-Realtime-2.1} (Azure deployment~\cite{azurerealtimeaudio}) & \texttt{session.\allowbreak{}update}
  & 24\,/\,24\,kHz \\
MOSS & \texttt{moss-live-1.0} & \texttt{attach\_\allowbreak{}session}, \texttt{media.\allowbreak{}mode},
  \texttt{media.\allowbreak{}configure}, \texttt{session.\allowbreak{}update} & 16\,/\,48\,kHz \\
Gemini Live & \texttt{gemini-3.8-live} & \texttt{setup} & 16\,/\,24\,kHz \\
\bottomrule
\end{tabularx}
\end{table}

Qwen Omni and GPT Realtime use a server VAD threshold of 0.5 and a 500\,ms silence setting in the server VAD configuration. The full-duplex results also include the semantic modes listed in Section~\ref{sec:experiments}. Doubao and MOSS use built-in turn detection. Qwen Omni and GPT Realtime expose configurable interruption handling. Doubao requires video to be declared in the session configuration, and Qwen Omni drops frames sent during generation. Sampling follows provider defaults.

\paragraph{Gemini Live adapter.}
Gemini Live uses server-side activity detection in autonomous replay. The adapter derives response starts from the first 	exttt{modelTurn} message, reads audio from inline data, and records cancellation from the server's 	exttt{interrupted} flag. Transcription-only messages do not start a response. Full-duplex settings use 20,ms audio chunks, a 30,ms scheduling tick, a 500,ms silence setting, and an 8,s output tail.

\paragraph{Task-specific delivery.}
Table~\ref{tab:task-delivery} summarizes delivery and output settings. After media delivery, the standard wind-down uploads silence. A session is busy during generation or during the first 3 s after the last reported user speech stop if a new reply has yet to start. Let $t_{\mathrm{last}}$ be the last downlink timestamp and $t_{\mathrm{end}}$ the end of the media. The wind-down ends at the earliest time satisfying
\begin{equation}
(\neg\operatorname{busy}(t)\land t-t_{\mathrm{last}}\geq0.7\,\mathrm{s})
\lor (t-t_{\mathrm{end}}\geq30\,\mathrm{s}).
\label{eq:settle}
\end{equation}
 Full-duplex runs use provider-specific wind-down settings. The harness reconstructs the playback track at the system's output sample rate, pads buffer underruns with zeros, and discards pending audio on a cancellation according to the provider's response identifiers. It then aligns and resamples the track for scoring. For delegated reception, replay advances according to the agent's questions and supplies the relevant recorded answer segments.

\begin{table}[htbp]
\centering\footnotesize
\caption{Delivery parameters by task. All tasks send frames at 2\,fps. Output
modality and barge-in handling are listed for Omni and GPT, the two systems on
which they are configurable.}
\label{tab:task-delivery}
\begin{tabularx}{\linewidth}{@{}lYllYYY@{}}
\toprule
Task & Delivery unit & $\Delta_c$ & $\Delta_b$ & Wind-down & Omni/GPT output; barge-in
  & Tool \\
\midrule
Full-duplex & Whole session & 20/100\,ms & -- & Fixed $T_{\mathrm{flush}}$
  & Audio; server stop & -- \\
Perception & Whole session & 20\,ms & 3000\,ms & Eq.~\eqref{eq:settle}
  & Text; server stop & \texttt{report\_\allowbreak{}current\_\allowbreak{}speaker} \\
Proactive & Whole session, optional cue & 20\,ms & 1500\,ms & Eq.~\eqref{eq:settle}
  & Text; client cancel & -- \\
Social norms & Whole session, optional cue & 20\,ms & 1500\,ms & Eq.~\eqref{eq:settle}
  & Text; client cancel & -- \\
Preset task & Gated segments & 20\,ms & 3000\,ms & Eq.~\eqref{eq:settle} per segment
  & Text; client cancel & -- \\
\bottomrule
\end{tabularx}
\end{table}

\begin{table}[htbp]
\centering\footnotesize
\caption{Request parameters for full-duplex timing.}
\label{tab:t1-params}
\begin{tabularx}{\linewidth}{@{}>{\raggedright\arraybackslash}p{0.17\linewidth}YYYY@{}}
\toprule
Parameter & Omni & Doubao & GPT & MOSS \\
\midrule
Output modality & \texttt{modalities=\allowbreak{}[text,\allowbreak{} audio]} & audio
  & \texttt{output\_\allowbreak{}modalities=\allowbreak{}[audio]} & audio \\
Voice & \texttt{Ethan} & \texttt{zh\_\allowbreak{}female\_\allowbreak{}vv\_\allowbreak{}jupiter\_\allowbreak{}bigtts} & \texttt{marin}
  & not configurable \\
Chunk $\Delta_c$ & 100\,ms & 20\,ms & 100\,ms & 100\,ms \\
Turn detection & \texttt{server\_\allowbreak{}vad} (0.5 / 500\,ms)
  & built-in, plus one commit after input & \texttt{server\_\allowbreak{}vad} (0.5 / 500\,ms)
  & built-in \\
\texttt{interrupt\_\allowbreak{}response} & \texttt{true} & -- & \texttt{true} & -- \\
Uplink after input ends & stop & commit, then silence & silence & silence \\
Wind-down $T_{\mathrm{flush}}$ & 1.0\,s & 8.0\,s & 8.0\,s & 1.0\,s \\
Video declaration & -- & \texttt{extension.\allowbreak{}video} (JPEG, 2\,fps) & --
  & \texttt{interval\_\allowbreak{}ms=\allowbreak{}500} \\
\bottomrule
\end{tabularx}
\end{table}

\paragraph{Judging and measurement.}
Full-duplex word timing uses \texttt{paraformer-v2}. Simultaneous interpreting uses \texttt{gpt-oss-120b}. The archived contextual-constraint runs used \texttt{qwen3.7-max} on a different scoring scale. The judge for the 0--3 rescores reported in Section~\ref{sec:experiments} is pending verification. Task-specific scoring units and missing-response handling are defined in Appendix~\ref{app:metric-definitions}.
\FloatBarrier

\subsection{Audio Front End for Speaker Perception}
\label{app:speaker-audio-front-end}
For each of the 130 speaker-perception recordings in Table~\ref{tab:speaker-results}, we replay the same video twice to each dialogue system. In the raw condition, the audio is the original 16\,kHz mixture from microphone channel~1. In the denoised condition, we use that same mixture to prepare one time-aligned speech track offline. During each replay, the dialogue system reports a speaker ID, an appearance description, and its own transcript of the utterance with each reply. We compare the reported ID with the annotated main speaker for that turn and compute the session-level character error rate from the system's transcripts and the corresponding reference utterances (Appendix~\ref{sec:speaker-scoring}).

\paragraph{Offline target selection.}
The front end follows one of the two scripted participants at a time. It starts with the first frame in which only one participant shows speaking activity. If no such frame occurs, it selects the track with the higher initial activity score, breaking a tie in favor of the first track. It holds the current target through silence and overlapping speech. A switch occurs on the third consecutive frame in which only the other participant is active. Thus, if B begins speaking over A, extraction continues to follow A until B has spoken alone for three frames. This rule selects the visual input for extraction. The reference-speaker rule for scoring system reports is given in Appendix~\ref{sec:speaker-scoring}.

\paragraph{Face tracking and speaking activity.}
We align the rectified 25\,fps video with channel~1, detect faces with S3FD~\cite{zhang2017s3fd}, and link detections across frames. MediaPipe mouth-corner landmarks~\cite{lugaresi2019mediapipe} locate a $96\times96$ grayscale mouth crop on each track. TalkNet~\cite{tao2021talknet}, with TalkSet weights, scores speaking activity from each face track and the shared audio. We smooth the scores and use lip motion relative to the surrounding face as an additional cue during audible speech. Motion on one track can identify a missed speaker when both scores are low, while motion on both tracks flags possible overlap. Figure~\ref{fig:a013-target-selection} shows the mouth crop selected for extraction in a restaurant recording.

\noindent\begin{minipage}{\linewidth}
  \centering
  \expandafter\def\csname @captype\endcsname{figure}
  \includegraphics[width=0.58\linewidth]{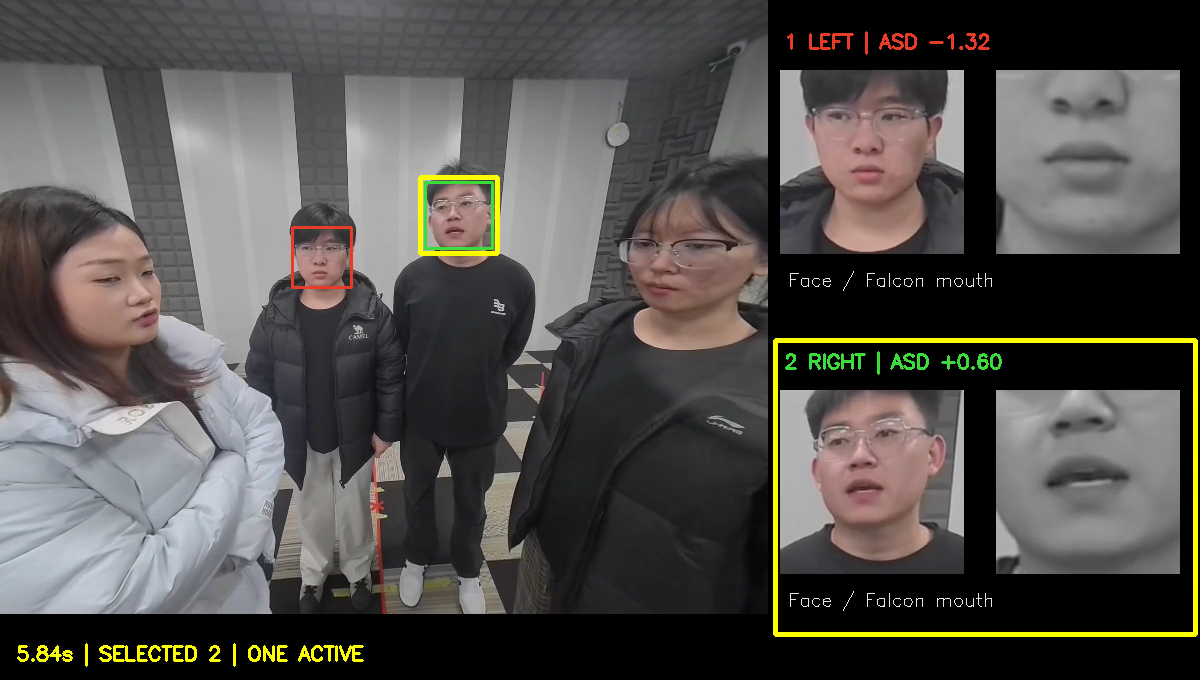}
  \caption{Offline target selection in the restaurant scene (A013, 5.84\,s). The robot view shows both participants. The face and mouth crops on the right correspond to participant~1 (top) and participant~2 (bottom). Only participant~2 is detected as speaking at this point. Yellow outlines mark the selected participant.}
  \label{fig:a013-target-selection}
\end{minipage}\par

\paragraph{Single-target extraction and mouth occlusion.}
For each interval with a fixed target, the small Falcon model~\cite{sang2026falcon} receives the channel~1 mixture and that participant's mouth sequence. We extend both inputs by up to 0.5\,s on either side for context, following the same face throughout. We retain only the output samples within the original interval and place them at their original times to produce a track of the same length as the mixture.

We use visual cues to automatically flag intervals in which the target speaker's mouth may be occluded. We use MediaPipe HandLandmarker~\cite{zhang2020mediapipehands} to detect overlap between a hand and the mouth region, and color cues within that region to flag other possible occlusions. During confirmed occlusions, we replace the Falcon output with the time-aligned output of GTCRN~\cite{rong2024gtcrn}, applied independently to the channel~1 mixture, with a short crossfade at each boundary. The selected identity stays fixed during substitution. Competing speech can remain audible because GTCRN receives only the microphone mixture.
\FloatBarrier

\end{document}